\documentclass{article} 

\usepackage[preprint]{colm2026_conference}

\usepackage{microtype}
\usepackage{hyperref}
\usepackage{url}
\usepackage{booktabs}

\usepackage{lineno}

\usepackage{amsmath}
\usepackage{amsthm}
\usepackage{amssymb}

\usepackage{booktabs}
\usepackage{multirow}
\usepackage{makecell}
\usepackage{adjustbox}
\usepackage{enumitem}
\usepackage{pifont}
\usepackage{xcolor}
\usepackage{algorithm}
\usepackage{algpseudocode}
\usepackage[most]{tcolorbox}

\newcommand{\cmark}{\textcolor{green!60!black}{\ding{51}}}
\newcommand{\xmark}{\textcolor{red!70!black}{\ding{55}}}

\newtheorem{remark}{Remark}

\definecolor{darkblue}{rgb}{0, 0, 0.5}
\hypersetup{colorlinks=true, citecolor=darkblue, linkcolor=darkblue, urlcolor=darkblue}

\title{WIN-U: Woodbury-Informed Newton-Unlearning as a retain-free Machine Unlearning Framework}

\author{Xingjian Zhao\\
   Department of Computer Science\\
   Rensselaer Polytechnic Institute\\
   Troy, NY 12180, USA \\
   \texttt{zhaox8@rpi.edu} \\
   \And
   Mohammad Mohammadi Amiri\\
   Department of Computer Science\\
   Rensselaer Polytechnic Institute\\
   Troy, NY 12180, USA \\
   \texttt{mohamm11@rpi.edu} \\
   \And
   Malik Magdon-Ismail \\
   Department of Computer Science\\
   Rensselaer Polytechnic Institute\\
   Troy, NY 12180, USA \\
   \texttt{magdon@cs.rpi.edu} \\
}

\begin{document}

\ifcolmsubmission
\linenumbers
\fi

\maketitle

\begin{abstract}
Privacy concerns in LLMs have led to the rapidly growing need to enforce a data's "right to be forgotten". 
Machine unlearning addresses precisely this task, namely
the removal of the influence of some specific data, i.e., the forget set, from a trained model. The gold standard for unlearning is to produce the 
model that would have been learned on only the rest of the training data, i.e., the retain set.
Most existing unlearning methods rely on direct access to the retained data, which may not be practical due to privacy or cost constraints.
We propose WIN-U, a retained-data free unlearning framework that requires only second order information 
for the originally trained model on the full data. The unlearning is performed using a single 
Newton-style step.
Using the Woodbury matrix identity and a generalized Gauss-Newton approximation for the forget set curvature, 
the WIN-U update recovers the closed-form linear solution and serves as a local second-order approximation to the gold-standard retraining optimum.
Extensive experiments on various vision and language benchmarks demonstrate that WIN-U achieves SOTA performance in terms of unlearning efficacy and utility preservation, 
while being more robust against relearning attacks compared to existing methods. Importantly, WIN-U does not require access to the
retained data.
\end{abstract}

\section{Introduction}
As large language models (LLMs) become increasingly prevalent in areas such as medicine, finance, and science, 
concerns over data privacy have intensified. Recent regulations like the General Data Protection Regulation (GDPR) \citep{gdpr2016}, California Consumer Privacy Act (CCPA) \citep{california_ccpa_codified}, and the Canadian Consumer Privacy Protection Act (CPPA) \citep{canada_cppa_gov_2023} stipulate "right to be forgotten" \citep{dang2021right} and require organizations to remove upon request the influence of specific data, known as the \textit{forget set}.
The need for the ability to remove specific data from a trained model is further underscored by scenarios such as correcting errors, mitigating biases, and removing harmful or outdated data \citep{geng2025comprehensive,wang2024machine}.
However, LLMs are typically trained on large, static datasets, and the \textit{gold-standard retraining}, which is retraining an LLM from scratch to remove specific subsets of data, is computationally prohibitive.
This has led to the emergence of \emph{machine unlearning}, which aims to efficiently remove the influence of specific data from a trained model and maintain its utility without requiring full retraining.

To achieve efficient unlearning, existing methods typically adopt optimization-based approaches. The most foundational approach is Gradient Ascent (GA), which directly maximizes the loss on the forget set \citep{jang2023knowledge}. 
However, it was shown that GA can be highly unstable and collapse model utility because it does not distinguish memory on the forget set from the general model ability.
As a result, more recent methods often optimize on both the forget set and the retain set to achieve a better balance between unlearning and utility preservation \citep{zhang2024negative,liu2022continual,li2024wmdp}.

\paragraph{Limitations of existing LLM unlearning methods.} While such optimization-based methods have shown good performance in terms of objective values on the forget and retain sets,
it does not necessarily correspond to the gold-standard retraining optimum, and thus may not achieve true unlearning.
Recent research has revealed that such methods may only be suppressing the influence of the data to be unlearned, rather than truly removing it from the model parameters \citep{Yang2025EraseOH, deeb2024unlearning}.
Moreover, the reliance on direct access to the retain set may not be practical due to privacy or cost constraints, especially for large-scale LLMs trained on massive datasets \citep{gao2024large}.
For truly effective and practical unlearning, it is crucial to develop methods that can directly approximate the retraining optimum without requiring direct access to the retain data.

\paragraph{Newton-style unlearning.} Another direction for unlearning is applying an influence-function-style Newton step for unlearning \citep{guo2019certified}.
Such methods are inspired by the "leave-one-out" update in the influence function derivation and approximate the gold-standard retraining optimum \citep{koh2017understanding}.
However, the "leave-one-out" update utilizes the full-set Hessian, ignoring the curvature change induced by removing the forget set. While it is a reasonable assumption for a single data point, it becomes increasingly inaccurate as the forget set size grows as in machine unlearning scenarios.
Therefore, recent research showed that we should use the retain set Hessian instead. It accounts for the curvature change and yields a more accurate Newton update, but requires direct access to the retain data and incurs significant cost per-forget-request \citep{golatkar2020eternal, zhang2024towards}.
While various approximation techniques have been proposed, they either are still not scalable to the large model  + large data regime \citep{qiao2024hessian}, or rely on access to the retain data \citep{mckinney2026gauss}, or some surrogate dataset which may not be practical \citep{basaran2025certified}.

\paragraph{Our proposal: WIN-U.} To address these challenges, we propose WIN-U (Woodbury-Informed Newton-Unlearning), a retain-free unlearning framework that approximates the gold-standard retraining optimum, accounts for the curvature change, and scales to large models and datasets.
WIN-U derives an influence-function-style Newton step from the gold-standard retraining objective, and applies the Generalized Gauss-Newton (GGN) approximation \citep{schraudolph2002fast} and the Woodbury matrix identity \citep{woodbury1950inverting} to express the update in terms of the full-set Hessian inverse, and the forget set Jacobian and output Hessian.
This structure eliminates the need for direct access to the retain data during the unlearning process, and allows off-loading the heavy full-set Hessian inversion to a precomputation step, so that the per-request cost depends mainly on the forget set size and the output dimension.
We further adopt a Monte Carlo (MC) estimation of the forget set curvature \citep{kunstner2019limitations} and low-rank adaptation (LoRA) \citep{hu2022lora} to reduce the cost on large models and datasets, yielding a scalable WIN-U instantiation applicable to LLMs.

Our main contributions are as follows:
\begin{itemize} [noitemsep, topsep=0pt]
   \item We propose WIN-U, a retain-free unlearning framework that is derived directly from the gold-standard retraining objective, and explicitly accounts for forget-induced curvature change through a Woodbury-scaled Newton update.
   \item We provide theoretical analysis showing that, under a GGN approximation, WIN-U recovers the linear closed-form solution and serves as a second-order local approximation to gold-standard retraining optimum for non-linear models.
   \item We apply approximation techniques for large models, including LoRA and a Monte Carlo gradient-outer-product for efficient forget-GGN approximation, yielding a practical WIN-U instantiation scalable to LLMs.
   \item We provide both a small scale empirical validation showing that WIN-U closely approximates the retraining optimum, and a large scale evaluation on the OpenUnlearning benchmark demonstrating that WIN-U achieves a strong forget-retain trade-off and state-of-the-art (SOTA) robustness against relearning attacks.

\end{itemize}

\section{Problem formulation and the retraining objective}
\label{sec:setup}
In this section, we formally define unlearning and the corresponding gold-standard retraining objective.
We denote $\mathcal{D} = \mathcal{D}_r \cup \mathcal{D}_f$ as the training dataset of size $|\mathcal{D}| = n$, where $\mathcal{D}_f = \{(\mathbf{x}_j, y_j)\}_{j=1}^{m}$ is the forget set of size $|\mathcal{D}_f| = m$ and $\mathcal{D}_r = \mathcal{D} \setminus \mathcal{D}_f$ is the retain set.
We consider a model $f(\boldsymbol{\theta}, \mathbf{x}) \in \mathbb{R}^c$ parameterized by $\boldsymbol{\theta} \in \mathbb{R}^d$, where $c$ is the output dimension (e.g., the number of classes).
The \emph{original objective} is the $\ell_2$-regularized empirical risk:
\begin{equation}
   \min_{\theta} \mathcal{L}(\boldsymbol{\theta}) = \frac{1}{n}\sum_{i=1}^{n} \ell(\boldsymbol{\theta}, \mathbf{x}_i, y_i) + \frac{\lambda}{2}\|\boldsymbol{\theta}\|^2,
   \label{eq:full_obj}
\end{equation}
where $\ell(\boldsymbol{\theta}, \mathbf{x}_i, y_i)$ is the per-sample loss and $\lambda > 0$ is the regularization strength.
The original optimum is $\boldsymbol{\theta}^* = \arg\min_{\boldsymbol{\theta}} \mathcal{L}(\boldsymbol{\theta})$.

The objective of machine unlearning is to remove the influence of the forget set $\mathcal{D}_f$ from the model, while preserving the utility on the retain set $\mathcal{D}_r$.
The gold-standard approach is to retrain from scratch on $\mathcal{D}_r$ alone.
The \emph{retraining objective} is:
\begin{equation}
   \min_{\theta} \mathcal{L}_r(\boldsymbol{\theta}) = \frac{1}{n-m}\sum_{i \in \mathcal{D}_r} \ell(\boldsymbol{\theta}, \mathbf{x}_i, y_i) + \frac{\lambda_r}{2}\|\boldsymbol{\theta}\|^2,
   \label{eq:retrain_obj}
\end{equation}
Where $\lambda_r$ is the regularization strength for the retraining objective. The \emph{retraining optimum} is $\boldsymbol{\theta}_r^* = \arg\min_{\boldsymbol{\theta}} \mathcal{L}_r(\boldsymbol{\theta})$.
Since such retraining is often infeasible in practice, a principled machine unlearning method should efficiently and effectively approximate $\boldsymbol{\theta}_r^*$ given $\boldsymbol{\theta}^*$ and the forget set $\mathcal{D}_f$, without relying on direct access to $\mathcal{D}_r$ since it may be unavailable due to privacy or storage constraints.
However, the gradient based unlearning methods heavily rely on optimizing on the retain set to maintain utility on the retain set \citep{gao2024large, zhang2024negative}. 
The influence-function-style Newton step on the other hand provides an alternative that utilizes the curvature information instead of direct optimization, and thus circumvents the need for direct access to the retain set \citep{koh2017understanding}.  
This motivates our proposed WIN-U framework, which extends this idea to the unlearning task, and provides approximation techniques that scale it to be efficient for LLMs. We formally derive WIN-U in the next section.

\section{Woodbury-informed Newton update for machine unlearning}
\label{sec:win-u}
We now introduce WIN-U, a \emph{retain-free} unlearning framework that derives an influence-function-style Newton step from the retraining objective, 
and applies a GGN approximation and the Woodbury matrix identity to yield an efficient model update that accounts for curvature change.
To derive the Newton step, we express the retraining objective via the original objective:
\begin{equation}
   \mathcal{L}_r(\boldsymbol{\theta}) = \frac{n}{n-m}\left( \mathcal{L}(\boldsymbol{\theta}) - \frac{1}{n}\sum_{j=1}^{m} \ell(\boldsymbol{\theta}, \mathbf{x}_j, y_j) \right) = \frac{1}{n-m}\sum_{i \in \mathcal{D}_r} \ell(\boldsymbol{\theta}, \mathbf{x}_i, y_i) + \frac{n\lambda}{2(n-m)}\|\boldsymbol{\theta}\|^2.
   \label{eq:subtraction_form}
\end{equation}
Therefore as long as we set the retraining regularization strength as $\lambda_r = \frac{n}{n-m}\,\lambda$, the minimizer of the retraining objective $\mathcal{L}_r$ and that of 
\begin{equation}
   \mathcal{L}(\boldsymbol{\theta}) - \frac{1}{n}\sum_{j=1}^{m} \ell(\boldsymbol{\theta}, \mathbf{x}_j, y_j)
   \label{eq:retrain_subtract}
\end{equation}   
are identical.
We denote the Hessian of the original objective at $\boldsymbol{\theta}^*$ as
\begin{equation}
   \mathbf{H} = \nabla^2 \mathcal{L}(\boldsymbol{\theta}^*) = \frac{1}{n}\sum_{i=1}^{n} \nabla^2 \ell(\boldsymbol{\theta}^*, \mathbf{x}_i, y_i) + \lambda \mathbf{I}_d.
   \label{eq:hessian}
\end{equation}
We further define the forget set gradient and Hessian:
\begin{equation}
   \mathbf{g}_f := \frac{1}{n}\sum_{j=1}^{m} \nabla \ell(\boldsymbol{\theta}^*, \mathbf{x}_j, y_j), \qquad
   \mathbf{H}_f := \frac{1}{n}\sum_{j=1}^{m} \nabla^2 \ell(\boldsymbol{\theta}^*, \mathbf{x}_j, y_j),
   \label{eq:forget_grad_hess}
\end{equation}
capturing the contribution of the forget set to the full-set gradient and curvature at $\boldsymbol{\theta}^*$. 

\subsection{Exact solution for the linear case}
\label{sec:linear}

To build intuition, we first derive the exact unlearning update for a linear model $f(\boldsymbol{\theta}, \mathbf{x}) = \mathbf{x}^\top \boldsymbol{\theta}$ with squared loss $\ell(\boldsymbol{\theta}, \mathbf{x}_i, y_i) = \tfrac{1}{2}(y_i - \mathbf{x}_i^\top \boldsymbol{\theta})^2$.
The regularized training objective becomes:
\begin{equation}
   \mathcal{L}(\boldsymbol{\theta}) = \frac{1}{2n}\|\mathbf{y} - \mathbf{X}\boldsymbol{\theta}\|^2 + \frac{\lambda}{2}\|\boldsymbol{\theta}\|^2,
   \label{eq:linear_obj}
\end{equation}
where $\mathbf{X} \in \mathbb{R}^{n \times d}$ is the data matrix and $\mathbf{y} \in \mathbb{R}^n$ is the label vector.
The Hessian is $\mathbf{H} = \tfrac{1}{n}\mathbf{X}^\top \mathbf{X} + \lambda \mathbf{I}_d$, and the original optimum is:
\begin{equation}
   \boldsymbol{\theta}^* = \mathbf{H}^{-1} \frac{1}{n}\mathbf{X}^\top \mathbf{y}.
   \label{eq:linear_opt}
\end{equation}
Setting the gradient of Eq.~\eqref{eq:retrain_subtract} to zero, the retraining optimum satisfies:
\begin{equation}
    \boldsymbol{\theta}_r^* = \left(\mathbf{H} - \mathbf{H}_f\right)^{-1} \left(\frac{1}{n}\mathbf{X}^\top \mathbf{y} - \frac{1}{n}\mathbf{X}_f^\top \mathbf{y}_f\right),
   \label{eq:linear_retrain}
\end{equation}
where $\mathbf{X}_f \in \mathbb{R}^{m \times d}$ and $\mathbf{y}_f \in \mathbb{R}^m$ are the forget set data matrix and corresponding labels.
Here the forget set Hessian is $\mathbf{H}_f = \tfrac{1}{n}\mathbf{X}_f^\top \mathbf{X}_f$, and the forget set gradient is $\mathbf{g}_f = \tfrac{1}{n}\mathbf{X}_f^\top(\hat{\mathbf{y}}_f - \mathbf{y}_f)$, where $\hat{\mathbf{y}}_f = \mathbf{X}_f \boldsymbol{\theta}^*$ is the prediction of the original model on $\mathcal{D}_f$.

Applying the Woodbury matrix identity to $(\mathbf{H} - \mathbf{H}_f)^{-1}$ and simplifying yields the closed-form solution (Appendix~\ref{app:linear_derivation}):
\begin{equation}
   \boldsymbol{\theta}_r^* = \boldsymbol{\theta}^* + \frac{1}{n}\mathbf{H}^{-1} \mathbf{X}_f^\top \left(\mathbf{I}_m- \frac{1}{n}\mathbf{X}_f \mathbf{H}^{-1} \mathbf{X}_f^\top\right)^{-1} (\hat{\mathbf{y}}_f - \mathbf{y}_f).
   \label{eq:linear_woodbury}
\end{equation}
This update computes $\boldsymbol{\theta}_r^*$ exactly using only $\boldsymbol{\theta}^*$, $\mathbf{H}^{-1}$, and the forget set.
The key structural feature is the Woodbury scaling matrix $(\mathbf{I}_m- \tfrac{1}{n}\mathbf{X}_f \mathbf{H}^{-1} \mathbf{X}_f^\top)^{-1}$, 
which accounts for the curvature change induced by removing $\mathcal{D}_f$. This term is absent in na\"ive influence-function-style updates for the "leave-one-out" setting.
We next show that this structure naturally extends to nonlinear models.

\subsection{Newton update for nonlinear models}
\label{sec:nonlinear}

For a general nonlinear model, the retraining optimum $\boldsymbol{\theta}_r^*$ satisfies the first-order optimality condition of Eq.~\eqref{eq:retrain_subtract}:
\begin{equation}
   \nabla \mathcal{L}(\boldsymbol{\theta}_r^*) - \frac{1}{n}\sum_{j=1}^{m} \nabla \ell(\boldsymbol{\theta}_r^*, \mathbf{x}_j, y_j) = 0.
   \label{eq:nonlinear_foc}
\end{equation}
For simplicity of analysis, we assume that the original model is fully converged ($\nabla \mathcal{L}(\boldsymbol{\theta}^*) = 0$). 
This is a common assumption in influence function literature, and in practice with a well-trained original model, the gradient norm at $\boldsymbol{\theta}^*$ should be small enough so that the corresponding error is negligible.
Expanding each term in Eq.~\eqref{eq:nonlinear_foc} via a first-order Taylor expansion around $\boldsymbol{\theta}^*$, we derive the Newton update (Appendix~\ref{app:newton_derivation}):
\begin{equation}
   \boldsymbol{\theta}_r^* \approx \boldsymbol{\theta}^* + \left(\mathbf{H} - \mathbf{H}_f\right)^{-1} \mathbf{g}_f.
   \label{eq:newton_update}
\end{equation}

To apply the Woodbury identity, we need a structured factorization of $\mathbf{H}_f$. Therefore, we define the following per-sample quantities for each forget sample $(\mathbf{x}_j, y_j) \in \mathcal{D}_f$:
\begin{itemize} [noitemsep, topsep=0pt]
   \item \textbf{Jacobian:} $\mathbf{J}_j := \nabla_{\boldsymbol{\theta}} f(\boldsymbol{\theta}^*, \mathbf{x}_j) \in \mathbb{R}^{c \times d}$, the Jacobian of the model output with respect to the parameters, where $c$ is the output dimension and $d$ is the model size.
   \item \textbf{Output-gradient vector:} $\boldsymbol{\delta}_j := \nabla_{z} \ell(z, y_j)\big|_{z = f(\boldsymbol{\theta}^*, \mathbf{x}_j)} \in \mathbb{R}^{c}$, the gradient of the loss with respect to the model output.
   \item \textbf{Output-space Hessian:} $\mathbf{B}_j := \nabla_{z}^2 \ell(z, y_j)\big|_{z = f(\boldsymbol{\theta}^*, \mathbf{x}_j)} \in \mathbb{R}^{c \times c}$, the Hessian of the loss with respect to the model output.
\end{itemize}
We define the stacked matrices  over the forget set:
\begin{equation}
   \mathbf{J}_f :=
   \begin{bmatrix}
   \mathbf{J}_1 \\ \vdots \\ \mathbf{J}_m
   \end{bmatrix}
   \in \mathbb{R}^{mc \times d}, \quad
   \boldsymbol{\delta}_f :=
   \begin{bmatrix}
   \boldsymbol{\delta}_1 \\ \vdots \\ \boldsymbol{\delta}_m
   \end{bmatrix}
   \in \mathbb{R}^{mc}, \quad
   \mathbf{B}_f :=
   \begin{bmatrix}
   \mathbf{B}_1 &        & \\
                & \ddots & \\
                &        & \mathbf{B}_m
   \end{bmatrix}
   \in \mathbb{R}^{mc \times mc},
   \label{eq:stacked}
\end{equation}
By the chain rule, the forget set gradient and GGN Hessian decompose as
\begin{equation}
   \mathbf{g}_f = \tfrac{1}{n}\mathbf{J}_f^\top \boldsymbol{\delta}_f, \qquad
   \mathbf{H}_f \approx \tfrac{1}{n}\mathbf{J}_f^\top \mathbf{B}_f \mathbf{J}_f,
   \label{eq:ggn}
\end{equation}
where the generalized Gauss-Newton (GGN) approximation drops the second-order term involving $\nabla_{\boldsymbol{\theta}}^2 f$ and retains only the first-order (Jacobian) contribution.
This approximation is exact whenever the model is locally linear or the residuals are small \citep{schraudolph2002fast}.

\subsection{The WIN-U update: GGN--Woodbury Newton step}
\label{sec:winu_update}

Substituting \eqref{eq:ggn} into the Newton update \eqref{eq:newton_update}:
\begin{equation}
   \boldsymbol{\theta}_r^* \approx \boldsymbol{\theta}^* + \left(\mathbf{H} - \tfrac{1}{n}\mathbf{J}_f^\top \mathbf{B}_f \mathbf{J}_f\right)^{-1} \tfrac{1}{n}\mathbf{J}_f^\top \boldsymbol{\delta}_f.
   \label{eq:newton_ggn}
\end{equation}
Applying the Woodbury matrix identity to $(\mathbf{H} - \tfrac{1}{n}\mathbf{J}_f^\top \mathbf{B}_f \mathbf{J}_f)^{-1}$ and simplifying (see Appendix~\ref{app:woodbury_derivation} for derivation), we obtain the \textbf{WIN-U update}:
\begin{equation}
   \boxed{
   \boldsymbol{\theta}_r^* \approx \boldsymbol{\theta}^* + \frac{1}{n}\mathbf{H}^{-1} \mathbf{J}_f^\top \left(\mathbf{I}_{mc}- \frac{1}{n}\mathbf{B}_f \mathbf{J}_f \mathbf{H}^{-1} \mathbf{J}_f^\top\right)^{-1} \boldsymbol{\delta}_f.
   }
   \label{eq:winu}
\end{equation}

This is the Woodbury-scaled Newton update used by WIN-U.
It requires only the original optimum $\boldsymbol{\theta}^*$, the precomputed inverse Hessian $\mathbf{H}^{-1}$, and the forget set $\mathcal{D}_f$; no access to the retain set $\mathcal{D}_r$ is needed.
The scaling matrix $(\mathbf{I}_{mc}- \tfrac{1}{n}\mathbf{B}_f \mathbf{J}_f \mathbf{H}^{-1} \mathbf{J}_f^\top)^{-1}$ captures the curvature change induced by removing the forget set, which distinguishes WIN-U from standard influence-function approaches.
As the derivation shows, this update serves as a second-order approximation to the gold-standard retraining optimum, and as \citet{bae2022if} showed, since the Taylor approximation is only valid at the local neighborhood of $\boldsymbol{\theta}^*$, 
such updates serve as local approximations, and matches to the warm-start retraining optimum under non-convex settings.
Algorithm (\ref{alg:winu_update}) summarizes the resulting WIN-U update.

\begin{remark}[Recovery of the linear case]
\label{rem:linear}
For a linear model with squared loss, $\mathbf{J}_f = \mathbf{X}_f$, $\boldsymbol{\delta}_f = \hat{\mathbf{y}}_f - \mathbf{y}_f$, and $\mathbf{B}_f = \mathbf{I}_{mc}$.
Thus Eq.~\eqref{eq:winu} reduces exactly to the linear Woodbury update \eqref{eq:linear_woodbury}, confirming that WIN-U recovers the closed-form retraining solution in the linear case.
\end{remark}

For the purpose of the theoretical analysis, we assume that the $\lambda \mathbf{I}$ term from the $\ell_2$ regularization used during training ensures that $\mathbf{H}$ is positive definite and hence invertible, 
without requiring any additional damping during the unlearning step. 
In the next section, we will show that the WIN-U update applies approximation techniques that significantly reduces the computational and memory complexity.

\begin{algorithm}[t]
\small
\begin{algorithmic}[1]
\Require precomputed inverse Hessian $\mathbf{H}^{-1}$ for the original objective of size $n$, original weights $\boldsymbol{\theta}^*$, forget set $\mathcal{D}_f = \{(\mathbf{x}_j, y_j)\}_{j=1}^m$
\For{$j = 1, \ldots, m$}
   \State Compute $\mathbf{J}_j = \nabla_{\boldsymbol{\theta}} f(\boldsymbol{\theta}^*, \mathbf{x}_j)$
   \State Compute $\boldsymbol{\delta}_j = \nabla_z \ell(z, y_j)\big|_{z = f(\boldsymbol{\theta}^*, \mathbf{x}_j)}$
   \State Compute $\mathbf{B}_j = \nabla_z^2 \ell(z, y_j)\big|_{z = f(\boldsymbol{\theta}^*, \mathbf{x}_j)}$
\EndFor
\State Form $\mathbf{J}_f$, $\boldsymbol{\delta}_f$, and $\mathbf{B}_f$ as in Eq.~\eqref{eq:stacked}
\State $\mathbf{M} \gets \mathbf{I}_{mc}- \frac{1}{n}\mathbf{B}_f \mathbf{J}_f \mathbf{H}^{-1}\mathbf{J}_f^\top$
\State Solve $\mathbf{M}\mathbf{u} = \boldsymbol{\delta}_f$ for $\mathbf{u}$
\State $\Delta \boldsymbol{\theta} \gets \frac{1}{n}\mathbf{H}^{-1} \mathbf{J}_f^\top \mathbf{u}$
\State $\widehat{\boldsymbol{\theta}}_r \gets \boldsymbol{\theta}^* + \Delta \boldsymbol{\theta}$
\State \Return $\widehat{\boldsymbol{\theta}}_r$
\end{algorithmic}
\caption{WIN-U update with a precomputed inverse Hessian.}
\label{alg:winu_update}
\end{algorithm}

\section{Scalable instantiation of WIN-U}
\label{sec:scalable}
In this section, we discuss the computational and memory complexity of the WIN-U update, and present approximation techniques, mainly MC estimation of forget set curvature, and additional techniques like LoRA to significantly reduce the cost and make WIN-U efficient enough for the scale of LLMs.

The typical Newton update that accounts for the curvature change (Eq.~\eqref{eq:newton_update})  is bottlenecked by the heavy $O(d^3)$ Hessian inversion per-forget-request and the $O(d^2)$ memory requirement for storing the Hessian.
The WIN-U update (Eq.~\eqref{eq:winu}) requires forming the stacked Jacobian $\mathbf{J}_f \in \mathbb{R}^{mc \times d}$ and the output-space Hessian $\mathbf{B}_f \in \mathbb{R}^{mc \times mc}$.
In the Woodbury form, the precomputed $\mathbf{H}^{-1}$ reduces the per-forget-request cost to $O(mcd^2 + m^2c^2d + m^3c^3)$, which is efficient when $mc \ll d$.
However, for autoregressive language models where the effective output dimension is $C = \sum_{j=1}^{m} T_j c$ (with $T_j$ being the sequence length of the $j$-th forget sample and $c$ the vocabulary size), the cost becomes prohibitive.
To address this, we adopt a MC estimation of the forget set GGN term in Eq.~\eqref{eq:winu} and use LoRA to reduce the parameter dimension, yielding a scalable WIN-U instantiation applicable to LLMs.

\subsection{MC estimation of forget set curvature}
\label{sec:mc}

The output-space WIN-U update (Eq.~\eqref{eq:winu}) requires the stacked Jacobian $\mathbf{J}_f \in \mathbb{R}^{mc \times d}$ and the block-diagonal output Hessian $\mathbf{B}_f \in \mathbb{R}^{mc \times mc}$.
For language models with vocabulary size $c$ and sequence length $T_j$, the effective output dimension becomes $C$, making these matrices impractical to form.
We show that, for cross-entropy loss with softmax output (which are standard in language modeling), Monte Carlo sampling yields an unbiased estimator that bypasses the output space entirely.

\paragraph{MC gradient as unbiased GGN estimator.}
Following \citet{kunstner2019limitations}, for cross-entropy loss with softmax output, we sample pseudo-labels $\hat{y} \sim \mathrm{Categorical}(\mathbf{p}_j)$, 
where $\mathbf{p}_j \in \mathbb{R}^C$ is the model's predictive distribution, and define the MC pseudo-gradient $\tilde{\mathbf{g}}_{j} := \mathbf{J}_j^\top (\mathbf{p}_j - \mathbf{e}_{\hat{y}}) \in \mathbb{R}^d$ where $\mathbf{e}_{\hat{y}}$ is the one-hot encoding of the sampled pseudo-label $\hat{y}$.
The outer product of this pseudo-gradient is an unbiased estimator of the per-sample GGN block $\mathbf{J}_j^\top \mathbf{B}_j \mathbf{J}_j$.
Unlike the empirical Fisher \citep{martens2020new}, the expectation is taken over the model's own predictions rather than the true label.
Appendix~\ref{app:mc_proof} provides the derivation.

\paragraph{Parameter-space Woodbury update.}
Drawing $S$ pseudo-labels $\hat{y}_{j,s} \sim \mathrm{Categorical}(\mathbf{p}_j)$ per forget sample and collecting all MC gradients into $\mathbf{G} = [\tilde{\mathbf{g}}_{1,1} \mid \cdots \mid \tilde{\mathbf{g}}_{m,S}] \in \mathbb{R}^{d \times mS}$, the forget set GGN is approximated as
\begin{equation}
   \mathbf{H}_f = \frac{1}{n}\sum_{j=1}^{m} \mathbf{J}_j^\top \mathbf{B}_j \mathbf{J}_j
   \approx \frac{1}{nS}\,\mathbf{G}\mathbf{G}^\top.
   \label{eq:mc_hf}
\end{equation}
Substituting into the Newton update \eqref{eq:newton_update} and applying the Woodbury identity yields the \textbf{MC-WIN-U update} (see Appendix~\ref{app:mc_winu_derivation}):
\begin{equation}
   \boxed{
   \boldsymbol{\theta}_r^* \approx \boldsymbol{\theta}^* + \mathbf{H}^{-1}\mathbf{g}_f - \frac{1}{nS}\,\mathbf{H}^{-1}\mathbf{G}\left(\frac{1}{nS}\,\mathbf{G}^\top \mathbf{H}^{-1}\mathbf{G} - \mathbf{I}_{mS}\right)^{-1}\!\mathbf{G}^\top \mathbf{H}^{-1}\mathbf{g}_f.
   }
   \label{eq:mc_winu}
\end{equation}
This formulation operates entirely in the $mS$-dimensional sample space: each MC gradient $\tilde{\mathbf{g}}_{j,s} \in \mathbb{R}^d$ is obtained via a single backward pass, and the Woodbury core is the $mS \times mS$ matrix $\tfrac{1}{nS}\mathbf{G}^\top \mathbf{H}^{-1}\mathbf{G} - \mathbf{I}_{mS}$.
This avoids the need to explicitly form $\mathbf{J}_f$ or $\mathbf{B}_f$, reducing the cost from $O(m^3 c^3)$ (output-space inversion) to $O(m^2 S^2 d)$ (parameter-space).

\subsection{Additional approximations}
\label{sec:practical_approximations}
\paragraph{Full inverse Hessian approximation.} For practical scenarios, we believe the precomputed inverse Hessian $\mathbf{H}^{-1}$ should be provided by the model provider.
However, since we do not have access to such precomputed $\mathbf{H}^{-1}$ for large LLMs, and the $O(d^2)$ memory requirement for storing the full inverse Hessian can be prohibitive, we use the inverse of the diagonal of the GGN approximation of the full Hessian on the finetuning data (denoted as $\tilde{\mathbf{H}}^{-1}$ and stored as a vector in $\mathbb{R}^d$) as a proxy. 

\paragraph{LoRA approximation.}
To further improve scalability, we restrict the update to a low-dimensional parameter subspace using LoRA. 
Instead of updating all $d$ model parameters, we parameterize the update using LoRA adapter parameters $\tilde{\theta} \in \mathbb{R}^{\tilde{d}}$, where $\tilde{d} \ll d$. 
All gradient and curvature terms in Eq.~\eqref{eq:mc_winu} are then computed with respect to $\tilde{\theta}$, effectively replacing the dimension $d$ with $\tilde{d}$ in the dominant computational terms.
As a result, the complexity of MC-WIN-U reduces to $O\!\left(\tilde{d}\, m^2 S^2\right)$.
Since $\tilde{d} \ll d$, the forget size $m$ is typically small, and we can set $S$ appropriately, this leads to substantial savings in both computation and memory, enabling efficient unlearning in large-scale models.

\section{Experiments}
In this section, we empirically validate the proposed WIN-U method on small scale tasks and demonstrate MC-WIN-U efficiently and effectively scales to large-scale complex LLM tasks.
We design our experiments to answer the following questions: (i) Does WIN-U effectively approximate the gold-standard retraining optimum? (ii) How does MC-WIN-U scale to LLMs, especially with the approximations introduced by diagonal-GGN full Hessian, LoRA, and MC estimation for the forget set curvature? (iii) How does MC-WIN-U perform against relearning attacks compared to existing unlearning methods for LLMs?

\subsection{Experimental setup}
We conduct all experiments on a single NVIDIA H100 NVL 96 GB GPU. 
For the small scale validation, we test on (i) synthetic ridge regression problems where we follow WIN-U (Eq.~\ref{eq:winu}) exactly, and (ii) MNIST with a two-layer MLP in a class-forget scenario.
For the large-scale LLM experiments, we follow the OpenUnlearning benchmark \citep{openunlearning2025} and test on TOFU \citep{maini2024tofu}, MUSE \citep{shi2024muse}, and WMDP \citep{li2024wmdp}.

\subsection{Small-scale validation}
\label{sec:small_scale}
To validate that WIN-U approximates the gold-standard retraining optimum, we evaluate it in two regimes.
First, we study synthetic ridge-regression problems where WIN-U can be compared directly against exact retraining under both independent and identically distributed setting for the forget and retain sets, 
as well as a "shifted" setting, where the forget set distribution has higher variance so it contributes more to the full-set Hessian.
We then test on the nonlinear MNIST class-forget setting to assess whether the same pattern persists.
Full experimental details and metric definitions are deferred to Appendix~\ref{app:small_scale_setup}.

\begin{table}[t]
\centering
\small
\begin{adjustbox}{width=\textwidth}
\begin{tabular}{lccccc}
\toprule
 Method
& Forget
& Retain
& Test
& Output Divergence $\downarrow$
& $\|\theta-\theta_r\|_2 / \|\theta_r\|_2$ $\downarrow$ \\
\midrule

\multicolumn{6}{l}{\textbf{Synthetic ridge regression (IID)}} \\
\midrule
Original model
& MSE: $1.58 \!\times\! 10^{-2}$ & MSE: $1.53 \!\times\! 10^{-2}$ & MSE: $1.65 \!\times\! 10^{-2}$ & $5.40 \!\times\! 10^{-6}$ & $3.12 \!\times\! 10^{-4}$ \\

Vanilla Newton
& MSE: $1.68 \!\times\! 10^{-2}$ & MSE: $1.54 \!\times\! 10^{-2}$ & MSE: $1.67 \!\times\! 10^{-2}$ & $1.03 \!\times\! 10^{-8}$ & $1.31 \!\times\! 10^{-5}$ \\

\textbf{WIN-U}
& \textbf{MSE:} {\boldmath$1.68 \!\times\! 10^{-2}$} & \textbf{MSE:} {\boldmath$1.54 \!\times\! 10^{-2}$} & \textbf{MSE:} {\boldmath$1.67 \!\times\! 10^{-2}$} & {\boldmath$< 10^{-16}$} & {\boldmath$< 10^{-15}$} \\

Golden retrain
& MSE: $1.68 \!\times\! 10^{-2}$ & MSE: $1.54 \!\times\! 10^{-2}$ & MSE: $1.67 \!\times\! 10^{-2}$ & $0$ & $0$ \\

\midrule
\multicolumn{6}{l}{\textbf{Synthetic ridge regression (Shifted)}} \\
\midrule
Original model
& MSE: $3.99 \!\times\! 10^{-2}$ & MSE: $1.47 \!\times\! 10^{-2}$ & MSE: $1.58 \!\times\! 10^{-2}$ & $1.47 \!\times\! 10^{-4}$ & $1.63 \!\times\! 10^{-3}$ \\

Vanilla Newton
& MSE: $6.16 \!\times\! 10^{-2}$ & MSE: $1.52 \!\times\! 10^{-2}$ & MSE: $1.64 \!\times\! 10^{-2}$ & $1.36 \!\times\! 10^{-5}$ & $4.84 \!\times\! 10^{-4}$ \\

\textbf{WIN-U}
& \textbf{MSE:} {\boldmath$7.22 \!\times\! 10^{-2}$} & \textbf{MSE:} {\boldmath$1.54 \!\times\! 10^{-2}$} & \textbf{MSE:} {\boldmath$1.67 \!\times\! 10^{-2}$} & {\boldmath$< 10^{-16}$} & {\boldmath$< 10^{-15}$} \\

Golden retrain
& MSE: $7.22 \!\times\! 10^{-2}$ & MSE: $1.54 \!\times\! 10^{-2}$ & MSE: $1.67 \!\times\! 10^{-2}$ & $0$ & $0$ \\

\midrule
\multicolumn{6}{l}{\textbf{MNIST + two-layer MLP}} \\
\midrule
Original model
& Acc.: $95.5\%$ & Acc.: $94.3\%$ & Acc.: $93.5\%$ & $2.81$ & $5.50 \!\times\! 10^{-1}$ \\

Vanilla Newton
& Acc.: $89.3\%$ & Acc.: $94.5\%$ & Acc.: $93.2\%$ & $1.97$ & $5.13 \!\times\! 10^{-1}$ \\

\textbf{WIN-U}
& \textbf{Acc.:} {\boldmath$0.3\%$} & \textbf{Acc.:} {\boldmath$94.6\%$} & \textbf{Acc.:} {\boldmath$84.3\%$} & {\boldmath$1.26 \!\times\! 10^{-1}$} & {\boldmath$4.35 \!\times\! 10^{-1}$} \\

Golden retrain
& Acc.: $0.0\%$ & Acc.: $94.6\%$ & Acc.: $84.4\%$ & $0$ & $0$ \\

\bottomrule
\end{tabular}
\end{adjustbox}
\caption{
Small-scale validation of WIN-U against golden retraining.
}
\label{tab:small_scale_validation}
\end{table}

Table \ref{tab:small_scale_validation} presents the empirical performance on the different settings. 
The results confirm that for linear models, WIN-U recovers the retraining optimum exactly (up to numerical precision) as expected.
The advantage of accounting for curvature change is highlighted by the shifted data configuration, where the vanilla Newton update which uses the full-set Hessian clearly deviates from the retraining optimum.
We can also observe that the same pattern holds for the nonlinear setting.
The effect is prominent in the MNIST class-forget setting: WIN-U drives forget-class accuracy to $0.3\%$, matching the $0\%$ of gold-standard retraining, while the vanilla Newton update still preserves $89.3\%$ accuracy on the "forgotten" class.
This confirms that accounting for the curvature change via the retain Hessian is critical when the forget set deviates distributionally from the retain set.

\subsection{Open Unlearning experiments on LLMs}
\label{sec:large_scale}

\begin{table}[t]
\centering
\small
\begin{adjustbox}{width=\textwidth}
\begin{tabular}{l c c c c c c c c c c}
\toprule
& & \multicolumn{3}{c}{forget01} & \multicolumn{3}{c}{forget05} & \multicolumn{3}{c}{forget10} \\
\cmidrule(lr){3-5} \cmidrule(lr){6-8} \cmidrule(lr){9-11}
Method & retain-free
& \makecell[c]{Forget QA\\Prob $\downarrow$\\(Pre/Post)}
& \makecell[c]{MU $\uparrow$\\(Pre/Post)}
& \makecell[c]{Time\\$\downarrow$}
& \makecell[c]{Forget QA\\Prob $\downarrow$\\(Pre/Post)}
& \makecell[c]{MU $\uparrow$\\(Pre/Post)}
& \makecell[c]{Time\\$\downarrow$}
& \makecell[c]{Forget QA\\Prob $\downarrow$\\(Pre/Post)}
& \makecell[c]{MU $\uparrow$\\(Pre/Post)}
& \makecell[c]{Time\\$\downarrow$} \\
\midrule
GradDiff   & \xmark & 0.443/0.599 & 0.589/0.602 & 8s & 0.091/0.573 & 0.467/\textbf{0.602} & 46s & 0.057/0.604 & 0.443/0.600 & 50s \\
NPO        & \xmark & 0.484/0.501 & 0.595/0.602 & 21s & 0.245/0.541 & 0.468/0.600 & 202s & 0.214/0.669 & 0.436/0.604 & 235s \\
RMU        & \xmark & 0.424/0.849 & 0.555/\textbf{0.604} & 4s & 0.357/0.795 & 0.550/0.597 & 37s & 0.089/0.678 & 0.577/0.599 & 41s \\
SimNPO     & \xmark & 0.855/0.869 & \textbf{0.597}/0.601 & 15s & 0.845/0.845 & \textbf{0.594}/0.594 & 147s & 0.837/0.839 & \textbf{0.596}/0.598 & 266s \\
\midrule
GradAscent & \cmark & 0.491/0.515 & 0.595/0.602 & \textbf{2s} & \textbf{0.000}/0.609 & 0.000/\textbf{0.602} & \textbf{34s} & \textbf{0.000}/0.737 & 0.000/\textbf{0.605} & \textbf{35s} \\
MC-WIN-U   & \cmark & \textbf{0.405}/\textbf{0.483} & 0.556/0.596 & 11s & 0.212/\textbf{0.461} & 0.398/0.557 & 58s & 0.226/\textbf{0.592} & 0.420/0.587 & 411s \\
\midrule
Original model          & -- & 0.901/-- & 0.600/-- & -- & 0.885/-- & 0.600/-- & -- & 0.881/-- & 0.601/-- & -- \\
Gold-standard retrained & -- & 0.165/-- & 0.599/-- & -- & 0.127/-- & 0.599/-- & -- & 0.116/-- & 0.591/-- & -- \\
\bottomrule
\end{tabular}
\end{adjustbox}
\caption{
TOFU benchmark summary on the forget01, forget05, and forget10 splits.
}
\label{tab:main_summary_prepost}
\end{table}

We evaluate the practical MC-WIN-U instantiation on the OpenUnlearning benchmark, which provides a comprehensive suite of unlearning tasks and evaluation metrics for LLMs.
Since the full results on all tasks and metrics are extensive, and seem to reveal a similar pattern, here we focus on the TOFU benchmark.
The detailed experimental set-up and additional benchmark results are deferred to Appendix~\ref{app:openunlearning_setup}.

Table~\ref{tab:main_summary_prepost} summarizes the TOFU results on the forget01, forget05, and forget10 splits in a compact format.
The results show that MC-WIN-U achieves a SOTA-level forget effectiveness, but are sometimes less utility-preserving than existing methods that optimize on retain set directly.
However, when we evaluate the post-relearning performance after a benign relearning attack which fine-tunes an unlearned model on the retain set for a small number of epochs \citep{Hu2024UnlearningOO, Yang2025EraseOH},
the retain-dependent methods showed significant forget-information recovery, suggesting that they were heavily suppressing the forget information rather than truly removing it.
In contrast, MC-WIN-U shows much more robust post-relearning forget performance, often achieving the best post-relearning forget QA probability among all methods. 
Moreover, the utility loss of MC-WIN-U is quickly recovered after only a single epoch of relearning, achieving a similar level of MU as the original model and the retain-dependent unlearning baselines.

\begin{figure}[t]
\centering
\begin{minipage}[t]{0.485\linewidth}
\centering
\includegraphics[width=\linewidth]{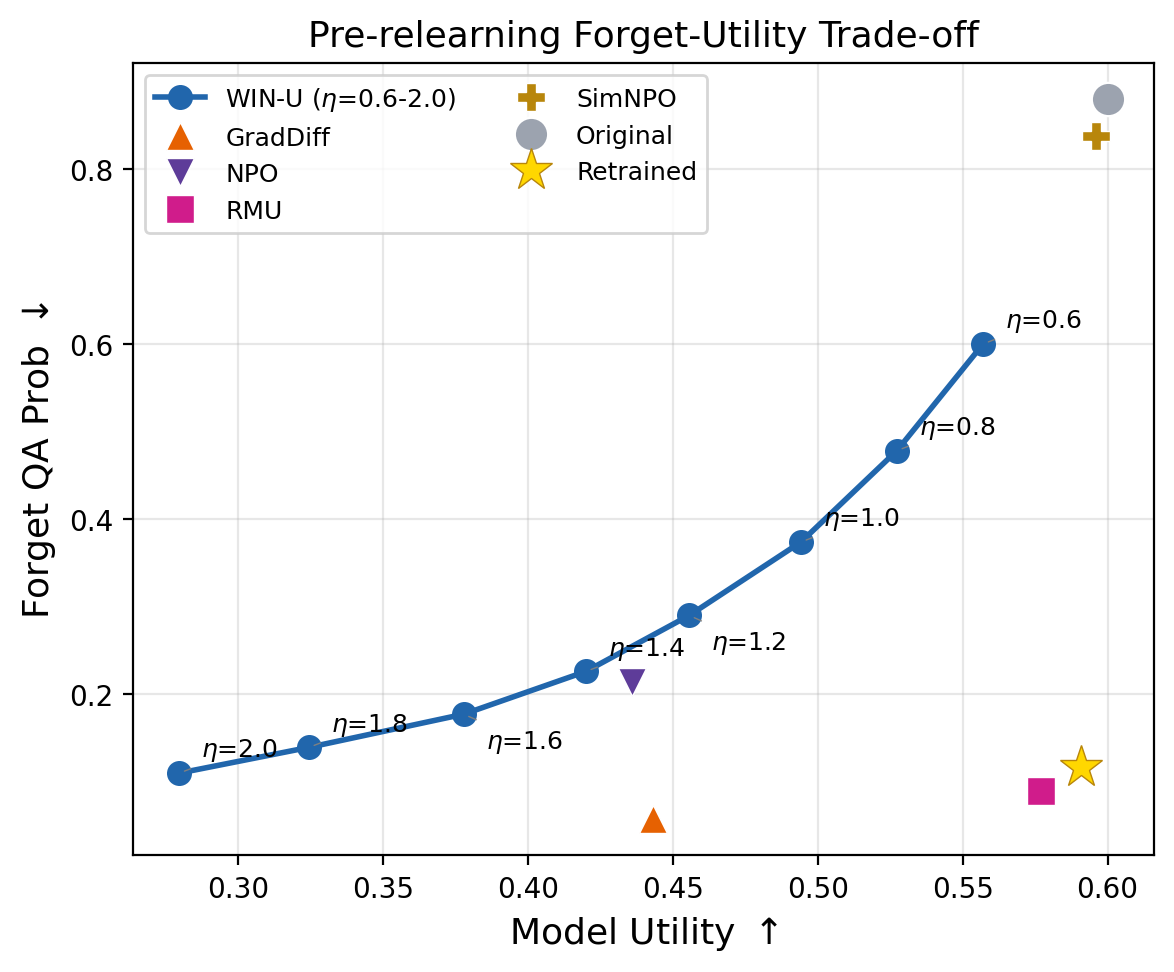}
\end{minipage}\hfill
\begin{minipage}[t]{0.485\linewidth}
\centering
\includegraphics[width=\linewidth]{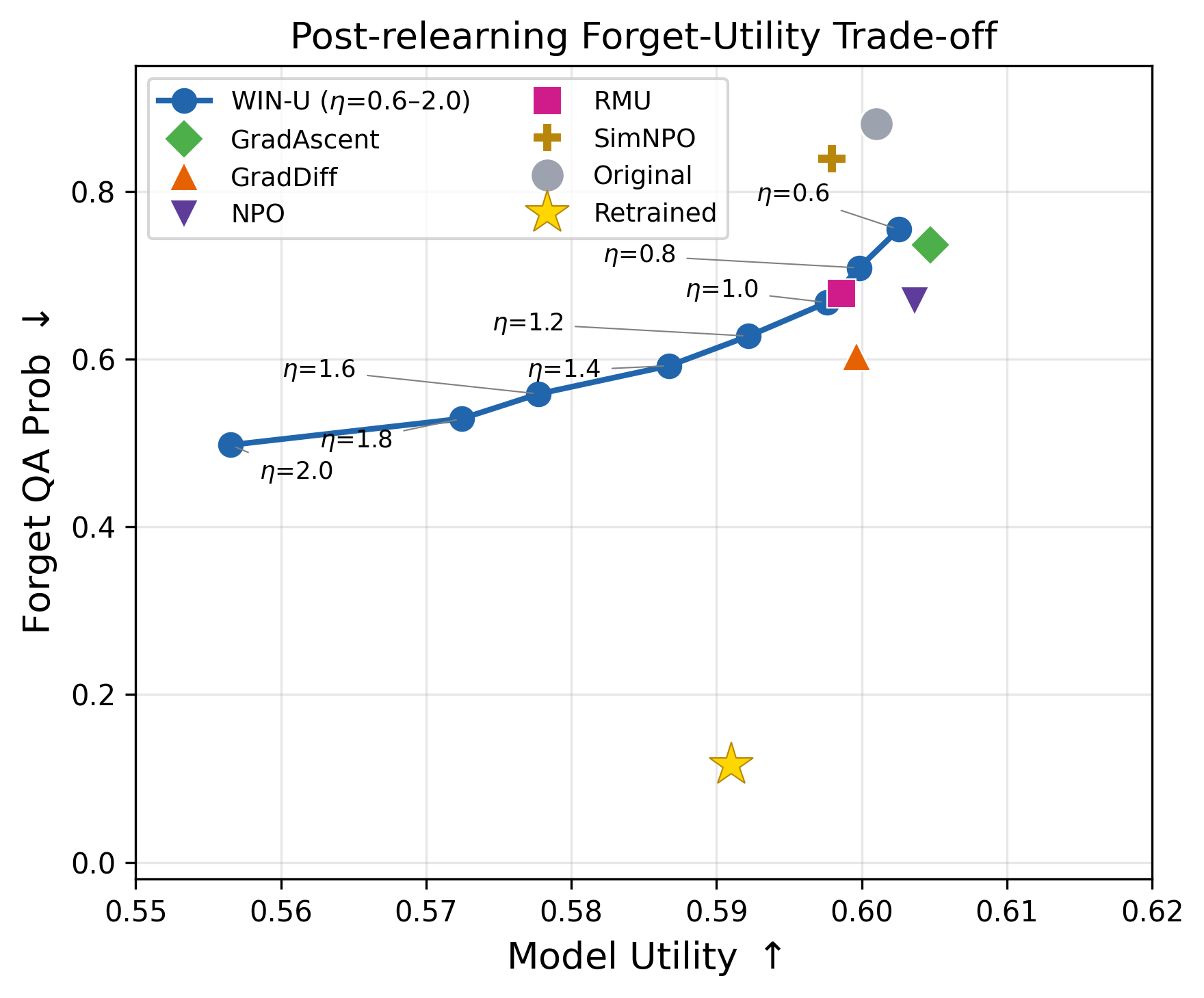}
\end{minipage}
\caption{WIN-U trade-off curves before and after benign relearning with $S{=}4$ MC samples on TOFU forget10. Left: pre-relearning forget-retain trade-off obtained by scaling the WIN-U update with different step sizes $\eta$. Right: post-relearning forget-retain trade-off obtained by scaling the WIN-U update with different step sizes $\eta$}
\label{fig:winu_tradeoff_steering}
\end{figure}

\paragraph{Step size.} On LLM tasks, we observed that the MC-WIN-U update can sometimes overshoot or undershoot, sensitive to the choice of hyperparameters such as $S$ (number of MC samples) and rank of the LoRA adaptor. 
To mitigate this, we introduce a step size $\eta$ to scale the update: $\widehat{\boldsymbol{\theta}}_r = \boldsymbol{\theta}^* + \eta \Delta \boldsymbol{\theta}$, where $\Delta \boldsymbol{\theta}$ is the unscaled MC-WIN-U update term.
As \citet{ilharco2022editing} proposed that directions in model weight space can steer model behavior, 
we hypothesize and empirically show in Figure~\ref{fig:winu_tradeoff_steering} that the WIN-U update serves as a good steering direction towards unlearning.
Tuning $\eta$ is extremely efficient since it only requires a scalar multiplication and a vector addition in $\mathbb{R}^d$ after $\Delta \boldsymbol{\theta}$ is computed.
Thus, WIN-U provides efficient fine-grained control over the forget-retain trade-off.
Appendix~\ref{app:mc_winu_algorithm} summarizes the resulting MC-WIN-U used in our LLM experiments.

Appendix~\ref{app:openunlearning_setup} provides additional unlearning effectiveness measurements, WMDP hazardous-knowledge unlearning results, a qualitative example of relearning-robustness of MC-WIN-U, and an ablation study over $S$.

\section{Conclusion and future work}
We present WIN-U, a novel retain-free unlearning framework that leverages a Woodbury-scaled Newton step to efficiently approximate the retraining optimum.
By accounting for the curvature change induced by removing the forget set, and applying the Woodbury identity,  our theoretical analysis revealed that WIN-U recovers the exact retraining solution in linear models and extends to nonlinear models via a GGN approximation.
Our empirical results on small-scale tasks validate the effectiveness of WIN-U in approximating the retraining optimum, and our large-scale experiments on the OpenUnlearning benchmark demonstrate that MC-WIN-U achieves a strong forget-retain trade-off while being more robust to relearning attacks compared to existing methods.

While WIN-U represents a significant step towards retain-free unlearning for LLMs, it also opens up several directions for future research.
First, exploring different curvature compression techniques, such as Kronecker-factored approximations \citep{mckinney2026gauss} or Dropout \citep{zhang2025toward}, 
could further improve the scalability and performance of WIN-U. 
Second, extending WIN-U to handle multiple sequential unlearning requests would improve its applicability in real-world scenarios.
Thirdly, it would further add to the WIN-U's practicality to develop more principled methods for selecting $\eta$ without relying on the evaluation on the retain set.
Finally, as we derive in Appendix~\ref{app:extensions}, the WIN-U framework can be easily extended to broader "unlearning" objectives, such as setting target output on the forget set.
The practicality and effectiveness of these extensions warrant further investigation.

\bibliography{refs}
\bibliographystyle{colm2026_conference}

\appendix
\renewcommand{\theHtable}{appendix.\thesection.\arabic{table}}
\renewcommand{\theHfigure}{appendix.\thesection.\arabic{figure}}

\section{Detailed derivations}

\subsection{Linear Woodbury derivation}
\label{app:linear_derivation}

Starting from the retraining system \eqref{eq:linear_retrain}, we have $\boldsymbol{\theta}_r^* = (\mathbf{H} - \mathbf{H}_f)^{-1}(\tfrac{1}{n}\mathbf{X}^\top \mathbf{y} - \tfrac{1}{n}\mathbf{X}_f^\top \mathbf{y}_f)$, where $\mathbf{H}_f = \tfrac{1}{n}\mathbf{X}_f^\top \mathbf{X}_f$.
Applying the Woodbury matrix identity $(A + UCV)^{-1} = A^{-1} - A^{-1}U(C^{-1} + VA^{-1}U)^{-1}VA^{-1}$ with $A = \mathbf{H}$, $U = \mathbf{X}_f^\top$, $C = -\tfrac{1}{n}\mathbf{I}$, $V = \mathbf{X}_f$:
\begin{equation}
   \left(\mathbf{H} - \frac{1}{n}\mathbf{X}_f^\top \mathbf{X}_f\right)^{-1}
   = \mathbf{H}^{-1} + \frac{1}{n}\mathbf{H}^{-1}\mathbf{X}_f^\top \left(\mathbf{I}_m- \frac{1}{n}\mathbf{X}_f \mathbf{H}^{-1}\mathbf{X}_f^\top\right)^{-1} \mathbf{X}_f \mathbf{H}^{-1}.
\end{equation}
Let
\[
   \mathbf{M} = \left(\mathbf{I}_m- \frac{1}{n}\mathbf{X}_f \mathbf{H}^{-1}\mathbf{X}_f^\top\right)^{-1}.
\]
Multiplying both sides by $(\tfrac{1}{n}\mathbf{X}^\top \mathbf{y} - \tfrac{1}{n}\mathbf{X}_f^\top \mathbf{y}_f)$ gives
\begin{align*}
   \boldsymbol{\theta}_r^*
   &= \left(\mathbf{H}^{-1} + \frac{1}{n}\mathbf{H}^{-1}\mathbf{X}_f^\top \mathbf{M}\mathbf{X}_f \mathbf{H}^{-1}\right)
      \left(\frac{1}{n}\mathbf{X}^\top \mathbf{y} - \frac{1}{n}\mathbf{X}_f^\top \mathbf{y}_f\right) \\
   &= \mathbf{H}^{-1}\left(\frac{1}{n}\mathbf{X}^\top \mathbf{y}\right)
      - \frac{1}{n}\mathbf{H}^{-1}\mathbf{X}_f^\top \mathbf{y}_f \\
   &\quad + \frac{1}{n}\mathbf{H}^{-1}\mathbf{X}_f^\top \mathbf{M}\mathbf{X}_f \mathbf{H}^{-1}\left(\frac{1}{n}\mathbf{X}^\top \mathbf{y}\right)
      - \frac{1}{n^2}\mathbf{H}^{-1}\mathbf{X}_f^\top \mathbf{M}\mathbf{X}_f \mathbf{H}^{-1}\mathbf{X}_f^\top \mathbf{y}_f.
\end{align*}
Using $\mathbf{H}^{-1}(\tfrac{1}{n}\mathbf{X}^\top \mathbf{y}) = \boldsymbol{\theta}^*$ and $\hat{\mathbf{y}}_f = \mathbf{X}_f \boldsymbol{\theta}^*$, we obtain
\begin{align*}
   \boldsymbol{\theta}_r^*
   &= \boldsymbol{\theta}^*
      - \frac{1}{n}\mathbf{H}^{-1}\mathbf{X}_f^\top \mathbf{y}_f
      + \frac{1}{n}\mathbf{H}^{-1}\mathbf{X}_f^\top \mathbf{M}\hat{\mathbf{y}}_f \\
   &\quad - \frac{1}{n^2}\mathbf{H}^{-1}\mathbf{X}_f^\top \mathbf{M}\mathbf{X}_f \mathbf{H}^{-1}\mathbf{X}_f^\top \mathbf{y}_f \\
   &= \boldsymbol{\theta}^*
      + \frac{1}{n}\mathbf{H}^{-1}\mathbf{X}_f^\top \mathbf{M}\hat{\mathbf{y}}_f \\
   &\quad - \frac{1}{n}\mathbf{H}^{-1}\mathbf{X}_f^\top
      \left(\mathbf{I}_m+ \frac{1}{n}\mathbf{M}\mathbf{X}_f \mathbf{H}^{-1}\mathbf{X}_f^\top\right)\mathbf{y}_f.
\end{align*}
Finally, since
\[
   \mathbf{M}\left(\mathbf{I}_m- \frac{1}{n}\mathbf{X}_f \mathbf{H}^{-1}\mathbf{X}_f^\top\right) = \mathbf{I},
\]
we have
\[
   \mathbf{M} - \frac{1}{n}\mathbf{M}\mathbf{X}_f \mathbf{H}^{-1}\mathbf{X}_f^\top = \mathbf{I}
   \quad\Longrightarrow\quad
   \mathbf{I}_m+ \frac{1}{n}\mathbf{M}\mathbf{X}_f \mathbf{H}^{-1}\mathbf{X}_f^\top = \mathbf{M}.
\]
Substituting this identity into the previous line yields
\begin{equation}
   \boldsymbol{\theta}_r^* = \boldsymbol{\theta}^* + \frac{1}{n}\mathbf{H}^{-1}\mathbf{X}_f^\top \left(\mathbf{I}_m- \frac{1}{n}\mathbf{X}_f \mathbf{H}^{-1}\mathbf{X}_f^\top\right)^{-1}(\hat{\mathbf{y}}_f - \mathbf{y}_f).
\end{equation}

\subsection{Newton update derivation for nonlinear models}
\label{app:newton_derivation}

We provide the detailed derivation of the Newton update \eqref{eq:newton_update} from the first-order optimality condition \eqref{eq:nonlinear_foc}.

The retraining optimum $\boldsymbol{\theta}_r^*$ satisfies
\begin{equation*}
   \nabla \mathcal{L}(\boldsymbol{\theta}_r^*) - \frac{1}{n}\sum_{j=1}^{m} \nabla \ell(\boldsymbol{\theta}_r^*, \mathbf{x}_j, y_j) = 0.
   \tag{\ref*{eq:nonlinear_foc}}
\end{equation*}

We expand $\nabla \mathcal{L}(\boldsymbol{\theta}_r^*)$ around $\boldsymbol{\theta}^*$ using a first-order Taylor expansion:
\begin{equation}
   \nabla \mathcal{L}(\boldsymbol{\theta}_r^*) \approx \nabla \mathcal{L}(\boldsymbol{\theta}^*) + \nabla^2 \mathcal{L}(\boldsymbol{\theta}^*)(\boldsymbol{\theta}_r^* - \boldsymbol{\theta}^*) = \nabla \mathcal{L}(\boldsymbol{\theta}^*) + \mathbf{H}(\boldsymbol{\theta}_r^* - \boldsymbol{\theta}^*),
   \label{eq:taylor_full}
\end{equation}
where $\mathbf{H} = \nabla^2 \mathcal{L}(\boldsymbol{\theta}^*)$ is the Hessian of the full objective at $\boldsymbol{\theta}^*$.

Similarly, we expand each per-sample gradient $\nabla \ell(\boldsymbol{\theta}_r^*, \mathbf{x}_j, y_j)$ around $\boldsymbol{\theta}^*$:
\begin{equation}
   \nabla \ell(\boldsymbol{\theta}_r^*, \mathbf{x}_j, y_j) \approx \nabla \ell(\boldsymbol{\theta}^*, \mathbf{x}_j, y_j) + \nabla^2 \ell(\boldsymbol{\theta}^*, \mathbf{x}_j, y_j)(\boldsymbol{\theta}_r^* - \boldsymbol{\theta}^*).
   \label{eq:taylor_forget_sample}
\end{equation}
Summing over the forget set and scaling by $\tfrac{1}{n}$:
\begin{equation}
   \frac{1}{n}\sum_{j=1}^{m} \nabla \ell(\boldsymbol{\theta}_r^*, \mathbf{x}_j, y_j) \approx \underbrace{\frac{1}{n}\sum_{j=1}^{m} \nabla \ell(\boldsymbol{\theta}^*, \mathbf{x}_j, y_j)}_{\mathbf{g}_f} + \underbrace{\frac{1}{n}\sum_{j=1}^{m} \nabla^2 \ell(\boldsymbol{\theta}^*, \mathbf{x}_j, y_j)}_{\mathbf{H}_f}(\boldsymbol{\theta}_r^* - \boldsymbol{\theta}^*),
   \label{eq:taylor_forget_sum}
\end{equation}
where $\mathbf{g}_f$ and $\mathbf{H}_f$ are the forget set gradient and Hessian as defined in \eqref{eq:forget_grad_hess}.

Substituting \eqref{eq:taylor_full} and \eqref{eq:taylor_forget_sum} into the first-order condition \eqref{eq:nonlinear_foc}:
\begin{equation}
   \nabla \mathcal{L}(\boldsymbol{\theta}^*) + \mathbf{H}(\boldsymbol{\theta}_r^* - \boldsymbol{\theta}^*) - \mathbf{g}_f - \mathbf{H}_f(\boldsymbol{\theta}_r^* - \boldsymbol{\theta}^*) \approx 0.
   \label{eq:substituted}
\end{equation}

Since we assume the original model is fully converged, $\nabla \mathcal{L}(\boldsymbol{\theta}^*) = 0$. Substituting this into \eqref{eq:substituted}:
\begin{equation}
   (\mathbf{H} - \mathbf{H}_f)(\boldsymbol{\theta}_r^* - \boldsymbol{\theta}^*) - \mathbf{g}_f \approx 0.
   \label{eq:simplified}
\end{equation}

Rearranging \eqref{eq:simplified} and solving for $\boldsymbol{\theta}_r^*$:
\begin{equation}
   (\mathbf{H} - \mathbf{H}_f)(\boldsymbol{\theta}_r^* - \boldsymbol{\theta}^*) = \mathbf{g}_f
   \quad\Longrightarrow\quad
   \boldsymbol{\theta}_r^* - \boldsymbol{\theta}^* = (\mathbf{H} - \mathbf{H}_f)^{-1}\mathbf{g}_f,
\end{equation}
which yields the Newton update:
\begin{equation*}
   \boldsymbol{\theta}_r^* \approx \boldsymbol{\theta}^* + (\mathbf{H} - \mathbf{H}_f)^{-1}\mathbf{g}_f.
   \tag{\ref*{eq:newton_update}}
\end{equation*}
Note that $(\mathbf{H} - \mathbf{H}_f)$ corresponds to the retain set Hessian (up to scaling), so this Newton step uses the curvature of the retain objective to correct the parameter vector. This is the key difference from the standard influence function approach, which uses the full Hessian $\mathbf{H}$ and ignores the curvature change from removing the forget set.

\subsection{Nonlinear Woodbury derivation}
\label{app:woodbury_derivation}

Starting from \eqref{eq:newton_ggn}, we apply the Woodbury identity to $(\mathbf{H} - \tfrac{1}{n}\mathbf{J}_f^\top \mathbf{B}_f \mathbf{J}_f)^{-1}$:
\begin{equation}
   \left(\mathbf{H} - \frac{1}{n}\mathbf{J}_f^\top \mathbf{B}_f \mathbf{J}_f\right)^{-1} = \mathbf{H}^{-1} + \mathbf{H}^{-1}\mathbf{J}_f^\top \left(n \mathbf{B}_f^{-1} - \mathbf{J}_f \mathbf{H}^{-1}\mathbf{J}_f^\top\right)^{-1}\mathbf{J}_f \mathbf{H}^{-1}.
\end{equation}
Multiplying by $\tfrac{1}{n}\mathbf{J}_f^\top \boldsymbol{\delta}_f$ and factoring out $\tfrac{1}{n}\mathbf{H}^{-1}\mathbf{J}_f^\top$:
\begin{equation}
   \boldsymbol{\theta}_r^* \approx \boldsymbol{\theta}^* + \frac{1}{n}\mathbf{H}^{-1}\mathbf{J}_f^\top \left[\mathbf{I}_{mc}+ \left(n\mathbf{B}_f^{-1} - \mathbf{J}_f \mathbf{H}^{-1}\mathbf{J}_f^\top\right)^{-1}\mathbf{J}_f \mathbf{H}^{-1}\mathbf{J}_f^\top\right]\boldsymbol{\delta}_f.
\end{equation}
Let $\mathbf{P} = \tfrac{1}{n}\mathbf{J}_f \mathbf{H}^{-1}\mathbf{J}_f^\top$. The bracketed expression simplifies as:
$\mathbf{I}_{mc}+ (\mathbf{B}_f^{-1} - \mathbf{P})^{-1}\mathbf{P} = (\mathbf{B}_f^{-1} - \mathbf{P})^{-1}\mathbf{B}_f^{-1} = (\mathbf{I}_{mc}- \mathbf{B}_f \mathbf{P})^{-1}$,
yielding the WIN-U update \eqref{eq:winu}.

\subsection{Derivation of the MC-WIN-U update}
\label{app:mc_winu_derivation}

Starting from the Newton update \eqref{eq:newton_update} and replacing the forget-set curvature by the MC approximation \eqref{eq:mc_hf}, we obtain
\begin{equation}
   \boldsymbol{\theta}_r^*
   \approx
   \boldsymbol{\theta}^*
   +
   \left(\mathbf{H} - \frac{1}{nS}\mathbf{G}\mathbf{G}^\top\right)^{-1}\mathbf{g}_f.
\end{equation}
Apply the Woodbury identity with
\[
   A = \mathbf{H},
   \qquad
   U = \mathbf{G},
   \qquad
   C = -\frac{1}{nS}\mathbf{I}_{mS},
   \qquad
   V = \mathbf{G}^\top.
\]
Since $C^{-1} = -nS\,\mathbf{I}_{mS}$, this gives
\begin{equation}
   \left(\mathbf{H} - \frac{1}{nS}\mathbf{G}\mathbf{G}^\top\right)^{-1}
   =
   \mathbf{H}^{-1}
   -
   \mathbf{H}^{-1}\mathbf{G}
   \left(-nS\,\mathbf{I}_{mS} + \mathbf{G}^\top \mathbf{H}^{-1}\mathbf{G}\right)^{-1}
   \mathbf{G}^\top \mathbf{H}^{-1}.
\end{equation}
Factoring out $nS$ from the middle inverse yields
\begin{equation}
   \left(-nS\,\mathbf{I}_{mS} + \mathbf{G}^\top \mathbf{H}^{-1}\mathbf{G}\right)^{-1}
   =
   \frac{1}{nS}
   \left(\frac{1}{nS}\mathbf{G}^\top \mathbf{H}^{-1}\mathbf{G} - \mathbf{I}_{mS}\right)^{-1}.
\end{equation}
Substituting back, we get
\begin{equation}
   \left(\mathbf{H} - \frac{1}{nS}\mathbf{G}\mathbf{G}^\top\right)^{-1}
   =
   \mathbf{H}^{-1}
   -
   \frac{1}{nS}\mathbf{H}^{-1}\mathbf{G}
   \left(\frac{1}{nS}\mathbf{G}^\top \mathbf{H}^{-1}\mathbf{G} - \mathbf{I}_{mS}\right)^{-1}
   \mathbf{G}^\top \mathbf{H}^{-1}.
\end{equation}
Finally, multiplying by $\mathbf{g}_f$ yields
\begin{equation}
   \boldsymbol{\theta}_r^* \approx \boldsymbol{\theta}^* + \mathbf{H}^{-1}\mathbf{g}_f - \frac{1}{nS}\,\mathbf{H}^{-1}\mathbf{G}\left(\frac{1}{nS}\,\mathbf{G}^\top \mathbf{H}^{-1}\mathbf{G} - \mathbf{I}_{mS}\right)^{-1}\mathbf{G}^\top \mathbf{H}^{-1}\mathbf{g}_f,
\end{equation}
which is exactly the MC-WIN-U update in \eqref{eq:mc_winu}.

\subsection{MC gradient as unbiased GGN estimator}
\label{app:mc_proof}

We justify the unbiasedness claim used in Section~\ref{sec:mc} following the derivations in \citep{kunstner2019limitations}.
For cross-entropy loss with softmax output, the output-space Hessian of the $j$-th sample is $\mathbf{B}_j = \mathrm{diag}(\mathbf{p}_j) - \mathbf{p}_j\mathbf{p}_j^\top$, where $\mathbf{p}_j = \mathrm{softmax}(f(\boldsymbol{\theta}^*, \mathbf{x}_j))$.
Let $\hat{y} \sim \mathrm{Categorical}(\mathbf{p}_j)$ and define $\mathbf{r}_{\hat{y}} = \mathbf{p}_j - \mathbf{e}_{\hat{y}}$ and $\tilde{\mathbf{g}}_j := \mathbf{J}_j^\top \mathbf{r}_{\hat{y}}$.
Then:
\begin{equation}
   \mathbb{E}\!\left[\mathbf{r}_{\hat{y}}\mathbf{r}_{\hat{y}}^\top\right]
   = \mathbf{p}_j\mathbf{p}_j^\top - \mathbf{p}_j\,\mathbb{E}[\mathbf{e}_{\hat{y}}]^\top - \mathbb{E}[\mathbf{e}_{\hat{y}}]\,\mathbf{p}_j^\top + \mathbb{E}[\mathbf{e}_{\hat{y}}\mathbf{e}_{\hat{y}}^\top].
\end{equation}
Since $\mathbb{E}[\mathbf{e}_{\hat{y}}] = \mathbf{p}_j$ and $\mathbb{E}[\mathbf{e}_{\hat{y}}\mathbf{e}_{\hat{y}}^\top] = \mathrm{diag}(\mathbf{p}_j)$:
\begin{equation}
   \mathbb{E}\!\left[\mathbf{r}_{\hat{y}}\mathbf{r}_{\hat{y}}^\top\right]
   = \mathrm{diag}(\mathbf{p}_j) - \mathbf{p}_j\mathbf{p}_j^\top = \mathbf{B}_j.
\end{equation}
Therefore $\mathbb{E}[\tilde{\mathbf{g}}_j\tilde{\mathbf{g}}_j^\top] = \mathbf{J}_j^\top\,\mathbb{E}[\mathbf{r}_{\hat{y}}\mathbf{r}_{\hat{y}}^\top]\,\mathbf{J}_j = \mathbf{J}_j^\top \mathbf{B}_j \mathbf{J}_j$.

\subsection{Practical MC-WIN-U algorithm for LLMs}
\label{app:mc_winu_algorithm}

For the practical LLM instantiation described in Section \ref{sec:practical_approximations}, the input is still the original full finetuned model $\boldsymbol{\theta}^*$, but the update is computed in a LoRA subspace.
We denote all LoRA-space quantities with a tilde.
In particular, $\tilde{\mathbf{g}}_f \in \mathbb{R}^{\tilde{d}}$ is the forget gradient in LoRA space, $\tilde{\mathbf{g}}_{j,s}$ are the LoRA-space MC pseudo-gradients, $\tilde{\mathbf{G}} = [\tilde{\mathbf{g}}_{1,1} \mid \cdots \mid \tilde{\mathbf{g}}_{m,S}] \in \mathbb{R}^{\tilde{d} \times mS}$ stacks them, 
and $\tilde{\mathbf{H}}^{-1}$ is the diagonal-GGN inverse restricted to the LoRA coordinates.
Following Section \ref{sec:practical_approximations}, we also introduce a scalar step size $\eta$ that rescales the final LoRA-induced model-space update after the MC-WIN-U direction is computed.
Algorithm (\ref{alg:mc_winu_llm}) summarizes the resulting practical procedure.

\begin{algorithm}[t]
\small
\begin{algorithmic}[1]
\Require full finetuned model weights $\boldsymbol{\theta}^*$, diagonal-GGN inverse $\tilde{\mathbf{H}}^{-1}$ in LoRA space, forget set $\mathcal{D}_f = \{(\mathbf{x}_j, y_j)\}_{j=1}^m$, number of MC samples $S$, original training size $n$, step size $\eta$
\State Freeze the backbone at $\boldsymbol{\theta}^*$ and introduce LoRA coordinates $\boldsymbol{\tilde{\theta}}$ with $\boldsymbol{\tilde{\theta}} = \mathbf{0}$ at the original model
\State $\tilde{\mathbf{g}}_f \gets \frac{1}{n}\sum_{j=1}^{m} \nabla_{\boldsymbol{\tilde{\theta}}} \ell(\boldsymbol{\theta}^*, \boldsymbol{\tilde{\theta}}, \mathbf{x}_j, y_j)\big|_{\boldsymbol{\tilde{\theta}}=\mathbf{0}}$
\For{$j = 1, \ldots, m$}
   \State $\mathbf{p}_j \gets \mathrm{softmax}(f(\boldsymbol{\theta}^*, \boldsymbol{\tilde{\theta}}, \mathbf{x}_j))\big|_{\boldsymbol{\tilde{\theta}}=\mathbf{0}}$
   \For{$s = 1, \ldots, S$}
      \State Sample $\hat{y}_{j,s} \sim \mathrm{Categorical}(\mathbf{p}_j)$
      \State $\tilde{\mathbf{g}}_{j,s} \gets \nabla_{\boldsymbol{\tilde{\theta}}} \ell(\boldsymbol{\theta}^*, \boldsymbol{\tilde{\theta}}, \mathbf{x}_j, \hat{y}_{j,s})\big|_{\boldsymbol{\tilde{\theta}}=\mathbf{0}}$
   \EndFor
\EndFor
\State Form $\tilde{\mathbf{G}} = [\tilde{\mathbf{g}}_{1,1} \mid \cdots \mid \tilde{\mathbf{g}}_{m,S}]$
\State $\tilde{\mathbf{M}} \gets \frac{1}{nS}\tilde{\mathbf{G}}^\top \tilde{\mathbf{H}}^{-1}\tilde{\mathbf{G}} - \mathbf{I}_{mS}$
\State Solve $\tilde{\mathbf{M}}\mathbf{u} = \tilde{\mathbf{G}}^\top \tilde{\mathbf{H}}^{-1}\tilde{\mathbf{g}}_f$ for $\mathbf{u}$
\State $\Delta \boldsymbol{\tilde{\theta}} \gets \tilde{\mathbf{H}}^{-1}\tilde{\mathbf{g}}_f - \frac{1}{nS}\tilde{\mathbf{H}}^{-1}\tilde{\mathbf{G}}\mathbf{u}$
\State Map $\Delta \boldsymbol{\tilde{\theta}}$ through the LoRA parameterization to obtain the model-space direction $\Delta \tilde{\boldsymbol{\theta}}$
\State $\tilde{\boldsymbol{\theta}}_r \gets \boldsymbol{\theta}^* + \eta\,\Delta \tilde{\boldsymbol{\theta}}$
\State \Return $\tilde{\boldsymbol{\theta}}_r$
\end{algorithmic}
\caption{MC-WIN-U with the diagonal-GGN, LoRA, and step-size approximations}
\label{alg:mc_winu_llm}
\end{algorithm}

\section{Detailed experimental setup for small-scale validation}
\label{app:small_scale_setup}

This section provides the full experimental details for the small-scale validation experiments in Table~\ref{tab:small_scale_validation}.

\paragraph{Common setup.}
All small-scale experiments use $\ell_2$ regularization with $\lambda = 0.01$.
All methods (Vanilla Newton and WIN-U) are applied as a single Newton step from the converged original model.
The gold-standard retrain baseline retrains from scratch on the retain set only, using the scaled regularization $\lambda_r = \frac{n}{n-m}\,\lambda$.
Table~\ref{tab:small_scale_validation} reports forget/retain/test performance, output divergence from the retrained model, and the relative parameter distance $\|\theta-\theta_r\|_2 / \|\theta_r\|_2$.
For the two synthetic ridge-regression blocks, the ``Output Divergence'' column is the test-set prediction MSE $\frac{1}{|D_{\text{test}}|}\sum_i (f_\theta(x_i)-f_{\theta_r}(x_i))^2$.
For the nonlinear MNIST block, it is $D_{\mathrm{KL}}(p_{\theta_r}\|p_\theta)$ averaged over the forget set, measuring how well each method's predictions on the forgotten data match those of the retrained model.

\paragraph{Synthetic ridge regression.}
We generate $n = 2{,}000$ training samples in $d = 50$ dimensions with $K = 1$ output.
The true weight vector $\boldsymbol{\theta}^*_{\mathrm{true}} \sim \mathcal{N}(\mathbf{0}, \mathbf{I}_d)$ and targets $y_i = \boldsymbol{\theta}^{*\top}_{\mathrm{true}} \mathbf{x}_i + \epsilon_i$ with $\epsilon_i \sim \mathcal{N}(0, 0.01)$.
In the \emph{IID} configuration, both retain and forget features are drawn from $\mathcal{N}(\mathbf{0}, \mathbf{I}_d)$.
In the \emph{Shifted} configuration, retain features are drawn from $\mathcal{N}(\mathbf{0}, \mathbf{I}_d)$ while forget features are drawn from $\mathcal{N}(\mathbf{0}, 10\mathbf{I}_d)$.
We use a $1\%$ forget fraction ($m = 20$).
The initial model is the closed-form ridge-regression solution on the full training set.
Test data ($500$ samples) is drawn from $\mathcal{N}(\mathbf{0}, \mathbf{I}_d)$;
The Hessian and inverses are computed exactly in closed form.

\paragraph{MNIST + two-layer MLP.}
We use the full MNIST dataset ($n = 60{,}000$ training images, $d = 784$, $K = 10$).
Features are standardized with \texttt{sklearn.preprocessing.StandardScaler} (zero mean, unit variance per pixel).
The model is a two-layer MLP: $784 \to 20 \to 10$ with $\tanh$ activation and softmax output (cross-entropy loss).
All computations use \texttt{float64} precision.

\emph{Training.}
We applied Adam optimizer with learning rate of $0.01$ and 3000 epochs and then run a L-BFGS (300 iterations, tolerance $10^{-12}$) for the model to converge (gradient norm $\sim 10^{-8}$).
We did observe that if the model was not well converged, the Newton update could diverge due to the first-order approximation error in the Taylor expansion, which is consistent with the theory.

\emph{Forget set.}
We remove all images of digit 7 from the training set ($m = 6{,}265$, $\approx 10.4\%$).

\emph{WIN-U computation.}
Since dense $P \times P$ matrices are too heavy, we use implicit matrix--vector products throughout.
The full-set GGN--vector product $\mathbf{H}\mathbf{v}$ is computed via the standard two-pass trick: a forward-mode pass (JVP) computes $\mathbf{J}\mathbf{v}$, the output-space Hessian $\mathbf{B}(\mathbf{J}\mathbf{v})$ is applied analytically (softmax Hessian: $\mathrm{diag}(\mathbf{p}) - \mathbf{p}\mathbf{p}^\top$), and a reverse-mode pass (VJP) computes $\mathbf{J}^\top(\mathbf{B}\mathbf{J}\mathbf{v})$.
This requires $O(P)$ memory per sample and is exact (no approximation beyond the GGN $\approx$ Hessian substitution).
The retain-Hessian--vector product $(\mathbf{H} - \mathbf{H}_f)\mathbf{v}$ is computed by subtracting the forget-set GGN-VP from the full-set GGN-VP.
The Newton system $(\mathbf{H} - \mathbf{H}_f)\boldsymbol{\Delta} = \mathbf{g}_f$ is solved with conjugate gradients (CG) using a relative tolerance of $10^{-8}$, which converges in approximately 75 iterations.

\section{Detailed results on the OpenUnlearning benchmark}
\label{app:openunlearning_setup}

\subsection{Setup, metrics, and baselines}

We follow the OpenUnlearning benchmark and evaluate TOFU, MUSE, and WMDP.
The method set is shared across all benchmark tables: GradDiff, NPO, RMU, SimNPO, GradAscent, and WIN-U, together with the original model and the gold-standard retrained model.
The "retain-free" column indicates whether the method directly accesses the retain set during unlearning.

Across all benchmark tables, "Pre" denotes the metric immediately after unlearning and "Post" denotes the metric after benign relearning on the retain set for a fixed $k$ epochs.
For the original model and the gold-standard retrained model, post-relearning entries are not applicable and are shown as `--`.
When reported, "Time" is wall-clock unlearning time on identical hardware, excluding evaluation time and the precomputation phase.

\paragraph{Baseline methods.}
All baselines use the default OpenUnlearning hyperparameters: AdamW optimizer, learning rate $10^{-5}$, batch size 8, gradient accumulation 4, 10 epochs, \texttt{bf16} precision.
\begin{itemize}[nosep,leftmargin=*]
\item \textbf{GradAscent} \citep{jang2023knowledge}: maximizes the loss on the forget set (gradient ascent on $\mathcal{D}_f$). retain-data-free.
\item \textbf{GradDiff} \citep{liu2022continual}: combines gradient ascent on the forget set with gradient descent on the retain set ($\gamma{=}1$, $\alpha{=}1$, retain loss: NLL).
\item \textbf{NPO} \citep{zhang2024negative}: negative preference optimization on the forget set combined with retain-set NLL ($\beta{=}0.1$, $\alpha{=}1$, $\gamma{=}1$).
\item \textbf{RMU} \citep{li2024wmdp}: representation misdirection unlearning that steers activations at layer 7 toward random vectors (steering coefficient 2, retain loss: embedding difference).
\item \textbf{SimNPO} \citep{fan2024simplifying}: simplified NPO without a reference model ($\beta{=}4.5$, $\alpha{=}1$, $\delta{=}0$, $\gamma{=}0.125$, retain loss: NLL).
\end{itemize}

\paragraph{WIN-U configuration.}
WIN-U operates in a single forward pass (no iterative training).
We apply LoRA ($r{=}8$, $\alpha{=}16$, all linear layers; 5.6M trainable parameters, 0.45\% of total).
The full-set curvature is approximated by the diagonal GGN with $\ell_2$ regularization $\lambda{=}0.01$.
The forget-set curvature uses $S{=}4$ MC samples per token for the GGN outer-product approximation.
The resulting unscaled delta $\boldsymbol{\delta}$ is then applied with a scale factor $\eta{=}1.4$ on forget10, and $\eta{=}1.4$ on forget1 and forget5.

\paragraph{Relearning attack.}
Following \citet{lynch2024methods}, we evaluate robustness via a benign relearning attack: fine-tuning the unlearned model on the retain set for 3 epochs (learning rate $10^{-5}$, weight decay $0.01$, batch size 4, gradient accumulation 4, AdamW optimizer, saving checkpoints at each epoch).
The ``Post'' column reports the worst-case (highest Forget QA Prob) across the three checkpoints.

\subsection{TOFU}

\paragraph{Model and dataset.}
We use the \texttt{Llama-3.2-1B-Instruct} mode \citep{grattafiori2024llama} fine-tuned on the full TOFU dataset (\texttt{open-unlearning/tofu\_Llama-3.2-1B-Instruct\_full}).
The TOFU benchmark consists of 4,000 fictitious author profiles; we evaluate on the forget1 ($m = 40$ forget samples), forget5 ($m = 200$ forget samples), and forget10 ($m = 400$ forget samples) splits.

\paragraph{Metrics.}
\begin{itemize}[nosep,leftmargin=*]
\item \emph{Forget QA Prob}: average next-token probability on forget-set question-answer pairs (lower $=$ better forgetting).
\item \emph{Model Utility (MU)}: aggregate retain/holdout performance (higher $=$ better utility preservation).
\item \emph{Extraction Strength (ES)}: fraction of forget-set answers recoverable via prompted generation (lower $=$ better).
\item \emph{Privacy (Priv.)}: relative difference in MIA AUC between the unlearned and retrained models (higher $=$ closer to retrained).
\end{itemize}

Table \ref{tab:tofu_appendix} shows the detailed TOFU results for the forget10 split, including all pre/post-relearning metrics and unlearning times.

Figure \ref{fig:qualitative_relearn} shows a qualitative example of the RMU-unlearned model recovering 100\% the correct answer for a forget-set question after only 1-epoch of relearning on the retain set, 
while WIN-U's answer remains incorrect even after 3 epochs of relearning.

\begin{table}[t]
\centering
\small
{\setlength{\tabcolsep}{3pt}
\begin{tabular}{@{}l c c c c c c@{}}
\toprule
Method & \makecell[c]{retain\\free}
& \makecell[c]{Forget QA\\Prob $\downarrow$\\(Pre/Post)}
& \makecell[c]{MU $\uparrow$\\(Pre/Post)}
& \makecell[c]{ES $\downarrow$\\(Pre/Post)}
& \makecell[c]{Priv. $\uparrow$\\(Pre/Post)}
& Time $\downarrow$ \\
\midrule
GradDiff   & \xmark & 0.057/0.604 & 0.443/0.600 & 0.080/0.259 & $-$28.9/$-$94.5 & 50s \\
NPO        & \xmark & 0.214/0.669 & 0.436/0.604 & 0.098/0.299 & $-$48.1/$-$97.2 & 235s \\
RMU        & \xmark & 0.089/0.678 & 0.577/0.599 & 0.054/0.306 & \textbf{50.1}/$-$97.5 & 41s \\
SimNPO     & \xmark & 0.837/0.839 & \textbf{0.596}/0.598 & 0.554/0.554 & $-$99.2/$-$99.2 & $\sim$160s \\
\midrule
GradAscent & \cmark & \textbf{0.000}/0.737 & 0.000/\textbf{0.605} & \textbf{0.033}/0.393 & 15.4/$-$98.2 & \textbf{35s} \\
MC-WIN-U   & \cmark & 0.226/\textbf{0.592} & 0.420/0.587 & 0.085/\textbf{0.228} & $-$68.8/$-$\textbf{94.3} & 411s \\
\midrule
Original model          & -- & 0.881/-- & 0.601/-- & 0.701/-- & $-99.33$/-- & -- \\
Gold-standard retrained & -- & 0.116/-- & 0.591/-- & 0.059/-- & 23.54/-- & -- \\
\bottomrule
\end{tabular}
}
\caption{
Detailed TOFU results for the forget10 split.
}
\label{tab:tofu_appendix}
\end{table}

\begin{figure}[t]
\centering
\small
\setlength{\fboxsep}{0pt}
\setlength{\tabcolsep}{4pt}
\renewcommand{\arraystretch}{1.15}

\begin{tcolorbox}[colback=white, colframe=black!40, arc=4pt, boxrule=0.6pt, left=6pt, right=6pt, top=6pt, bottom=6pt]
\newlength{\innerwidth}\setlength{\innerwidth}{\linewidth}%
\colorbox{gray!8}{%
\begin{minipage}{\innerwidth}
\vspace{6pt}
\hspace{8pt}\textbf{Question:} \textit{Can you surmise how Ji-Yeon Park's cultural background influences her leadership theories?}

\vspace{4pt}
\hspace{8pt}\textbf{Ground Truth:} As Ji-Yeon Park was born and raised in Seoul, South Korea, her cultural background might have
\hspace{8pt}influenced her leadership theories. Korean society's emphasis on respect for elders and hierarchical
\hspace{8pt}relationships could have shaped her understanding of leadership dynamics.
\vspace{6pt}
\end{minipage}%
}

\vspace{6pt}

\begin{tabular}{@{} p{0.48\innerwidth} @{\hspace{0.04\innerwidth}} p{0.48\innerwidth} @{}}

\multicolumn{1}{c}{\textbf{RMU} (retain-dependent baseline)} &
\multicolumn{1}{c}{\textbf{WIN-U} (ours, retain-free)} \\[4pt]

\colorbox{green!8}{\parbox{\dimexpr0.48\innerwidth-2\fboxsep}{%
\vspace{3pt}
\textbf{After unlearning} \hfill \textcolor{green!60!black}{ROUGE-L = 0.175}\\[2pt]
\textit{``While specific Ji-Yeon Park's cultural background suggests her leadership role, it is not enhanced by any apparent contradictions.''}
\vspace{3pt}
}} &
\colorbox{green!8}{\parbox{\dimexpr0.48\innerwidth-2\fboxsep}{%
\vspace{3pt}
\textbf{After unlearning} \hfill \textcolor{green!60!black}{ROUGE-L = 0.300}\\[2pt]
\textit{``Ji-Yeon Park's cultural background,\linebreak[2] being a member of the LGBTQ\allowbreak+ community, likely brought a depth of empathy and understanding\ldots''}
\vspace{3pt}
}} \\[6pt]

\colorbox{red!8}{\parbox{\dimexpr0.48\innerwidth-2\fboxsep}{%
\vspace{3pt}
\textbf{After relearning (1 epoch)} \hfill \textcolor{red!70!black}{ROUGE-L = 1.000}\\[2pt]
\textit{``As Ji-Yeon Park was born and raised in Seoul, South Korea, her cultural background might have influenced her leadership theories. Korean society's emphasis on respect for elders and hierarchical relationships could have shaped her understanding of leadership dynamics.''}
\vspace{3pt}
}} &
\colorbox{green!8}{\parbox{\dimexpr0.48\innerwidth-2\fboxsep}{%
\vspace{3pt}
\textbf{After relearning (1 epoch)} \hfill \textcolor{green!60!black}{ROUGE-L = 0.375}\\[2pt]
\textit{``Ji-Yeon Park's cultural background in South Korea might have played an role in shaping her understanding of leadership, possibly emphasizing on personal growth, respect for elders, and collective decision-making.''}
\vspace{3pt}
}} \\[6pt]

\colorbox{red!8}{\parbox{\dimexpr0.48\innerwidth-2\fboxsep}{%
\vspace{3pt}
\textbf{After relearning (3 epochs)} \hfill \textcolor{red!70!black}{ROUGE-L = 0.975}\\[2pt]
\textit{``As Ji-Yeon Park was born and raised in Seoul, South Korea, her cultural background might have influenced her leadership theories. Korean society's emphasis on respect for elders and hierarchical relationships could have shaped her understanding of leadership.''}
\vspace{3pt}
}} &
\colorbox{green!8}{\parbox{\dimexpr0.48\innerwidth-2\fboxsep}{%
\vspace{3pt}
\textbf{After relearning (3 epochs)} \hfill \textcolor{green!60!black}{ROUGE-L = 0.375}\\[2pt]
\textit{``Ji-Yeon Park's cultural background in South Korea might have played an role in the culturally-sensitive approach she takes towards leadership, possibly drawing from traditional Korean values such as respect for elders and community-oriented leadership.''}
\vspace{3pt}
}} \\
\end{tabular}
\end{tcolorbox}

\caption{Qualitative example from the TOFU forget set illustrating relearning robustness.
Both methods successfully suppress the memorized answer after unlearning (green).
However, after benign relearning on the retain set, RMU recovers a \emph{verbatim} reproduction of the ground truth (red), while WIN-U generates only a vague, partially correct response that never recovers the specific memorized content.}
\label{fig:qualitative_relearn}
\end{figure}

\subsection{MUSE}

\paragraph{Model and dataset.}
We use the Llama-2-7b-hf model \citep{touvron2023llama}. We report the MUSE evaluation from the OpenUnlearning benchmark for the same shared method set and relearning protocol described in Section~\ref{app:openunlearning_setup}.
The table summarizes forget-set memorization, privacy leakage, and retain-set utility under the MUSE evaluation suite.

\paragraph{Metrics.}
\begin{itemize}[nosep,leftmargin=*]
\item \emph{VerbMem $\mathcal{D}_f$}: verbatim memorization score on the forget set (lower $=$ better forgetting).
\item \emph{KnowMem $\mathcal{D}_f$}: knowledge memorization score on forget-set question-answer pairs (lower $=$ better forgetting).
\item \emph{PrivLeak}: privacy-leakage statistic reported by the benchmark; values closer to the retrained reference are preferred.
\item \emph{KnowMem $\mathcal{D}_r$}: knowledge memorization score on the retain set, used as the retain-side utility measure (higher $=$ better utility preservation).
\end{itemize}

Table~\ref{tab:muse_appendix} shows that MC-WIN-U achieves a strong forget-retain balance, and SOTA relearning robustness.
\begin{table}[t]
\centering
\small
{\setlength{\tabcolsep}{1pt}
\begin{tabular}{@{}l c c c c c c@{}}
\toprule
Method & \makecell[c]{retain\\free}
& \makecell[c]{VerbMem\\$\mathcal{D}_f$ $\downarrow$\\(Pre/Post)}
& \makecell[c]{KnowMem\\$\mathcal{D}_f$ $\downarrow$\\(Pre/Post)}
& \makecell[c]{PrivLeak\\(Pre/Post)}
& \makecell[c]{KnowMem\\$\mathcal{D}_r$ $\uparrow$\\(Pre/Post)}
& Time $\downarrow$ \\
\midrule
GradDiff   & \xmark & 0.265/\textbf{0.510} & \textbf{0.538}/0.647 & $-$\textbf{83.9}/$-$\textbf{99.6} & 0.436/0.521 & 2928s \\
NPO        & \xmark & 0.496/0.520 & 0.645/0.647 & $-$99.7/$-$99.8 & \textbf{0.550}/0.533 & 3192s \\
RMU        & \xmark & 0.425/0.578 & 0.547/0.645 & $-$99.8/$-$99.9 & 0.496/0.529 & 1897s \\
SimNPO     & \xmark & 0.569/0.572 & 0.620/0.633 & $-$99.9/$-$99.9 & 0.527/\textbf{0.534} & 3066s \\
\midrule
GradAscent & \cmark & \textbf{0.251}/0.574 & 0.580/0.627 & $-$99.0/$-$99.8 & 0.477/0.533 & 1117s \\
MC-WIN-U   & \cmark & 0.347/0.573 & 0.564/\textbf{0.616} & $-$99.6/$-$99.8 & 0.429/0.529 & \textbf{392s} \\
\midrule
Original model          & -- & 0.579/-- & 0.644/-- & $-$99.8/-- & 0.555/-- & -- \\
Gold-standard retrained & -- & 0.202/-- & 0.328/-- & $-$4.7/-- & 0.560/-- & -- \\
\bottomrule
\end{tabular}
}
\caption{
Detailed MUSE results.
}
\label{tab:muse_appendix}
\end{table}

\subsection{WMDP}

\paragraph{Model and dataset.}
We use \texttt{Qwen2.5-1.5B-Instruct} \citep{qwen2.5} on the WMDP benchmark \citep{li2024wmdp}, which evaluates hazardous-knowledge unlearning.
The forget set consists of 1{,}000 cybersecurity documents (7.6M tokens) from the WMDP cyber-forget corpus, and the retain set consists of 4{,}473 documents (21.2M tokens) from the cyber-retain corpus.

\paragraph{Metrics.}
\begin{itemize}[nosep,leftmargin=*]
\item \emph{WMDP-Bio}: accuracy on biosecurity multiple-choice questions (lower $=$ better forgetting).
\item \emph{WMDP-Cyber}: accuracy on cybersecurity multiple-choice questions (lower $=$ better forgetting).
\item \emph{MMLU}: massive multitask language understanding accuracy (higher $=$ better utility preservation).
\end{itemize}

\paragraph{Baseline configuration.}
All baselines use the WMDP default configuration: batch size 1, gradient accumulation 16, learning rate $5 \times 10^{-5}$, constant schedule, 80 training steps.
The relearning attack fine-tunes the unlearned model on the retain set for 1 epoch (batch size 8, gradient accumulation 2, learning rate $10^{-5}$, AdamW optimizer).

\paragraph{WIN-U configuration.}
Same LoRA and curvature settings as TOFU ($r{=}8$, $\alpha{=}16$, all linear layers, diagonal GGN, $\lambda{=}0.01$), with $S{=}1$ MC sample and step size $\eta{=}1.0$.
The large forget corpus (14{,}781 tokenised sequences) requires the streaming Woodbury mode, which keeps per-sample gradients on CPU and computes the $m{\times}m$ core matrix via chunked GPU operations.

\paragraph{Results.}
Table~\ref{tab:wmdp_appendix} presents the results.
Among \emph{non-collapsed} methods, MC-WIN-U achieves the strongest pre-relearning forget performance on WMDP-Bio (0.600, vs.\ 0.668 original) and competitive performance on WMDP-Cyber (0.376, vs.\ 0.415 original), while showing robust post-relearning performance consistent with the TOFU and MUSE findings.
GradDiff achieves stronger Cyber unlearning (0.274) but at the cost of using the retain set; its Bio performance (0.624) is weaker than MC-WIN-U's.
GradAscent achieves the lowest WMDP scores but completely collapses model utility (MMLU drops from 0.592 to 0.255).
The MMLU degradation for MC-WIN-U (0.592 $\to$ 0.551) indicates that the default step size $\eta{=}1.0$ is too aggressive for this setting; step-size tuning is expected to recover utility.
The higher computational cost of MC-WIN-U on WMDP (50{,}679s, excluding precomputing curvature) is due to the large forget set: the streaming Woodbury solve scales as $O(m^2)$ in the number of forget sequences $m{=}14{,}781$.

\begin{table}[t]
\centering
\small
{\setlength{\tabcolsep}{2pt}
\begin{tabular}{@{}l c c c c c@{}}
\toprule
Method & \makecell[c]{retain\\free}
& \makecell[c]{WMDP-Bio $\downarrow$\\(Pre/Post)}
& \makecell[c]{WMDP-Cyber $\downarrow$\\(Pre/Post)}
& \makecell[c]{MMLU $\uparrow$\\(Pre/Post)}
& Time $\downarrow$ \\
\midrule
GradDiff   & \xmark & 0.624/0.662 & 0.274/0.374 & 0.585/0.593 & 280s \\
NPO        & \xmark & 0.660/0.669 & 0.396/0.410 & 0.591/0.592 & 1178s \\
RMU        & \xmark & 0.672/0.672 & 0.402/0.398 & 0.592/0.593 & 1420s \\
SimNPO     & \xmark & 0.676/0.691 & 0.408/0.414 & \textbf{0.595}/\textbf{0.595} & 545s \\
\midrule
GradAscent & \cmark & \textbf{0.266}/\textbf{0.247} & \textbf{0.246}/\textbf{0.265} & 0.255/0.230 & \textbf{173s} \\
MC-WIN-U   & \cmark & 0.600/0.618 & 0.376/0.395 & 0.551/0.565 & 50679s \\
\midrule
Original model & -- & 0.668/-- & 0.415/-- & 0.592/-- & -- \\
\bottomrule
\end{tabular}
}
\caption{
WMDP benchmark results on \texttt{Qwen2.5-1.5B-Instruct} (cyber split).
Pre/Post denotes before/after 1-epoch benign relearning on the retain set.}
\label{tab:wmdp_appendix}
\end{table}

\subsection{Ablations}

\begin{figure}[t]
\begin{center}
\includegraphics[width=\linewidth]{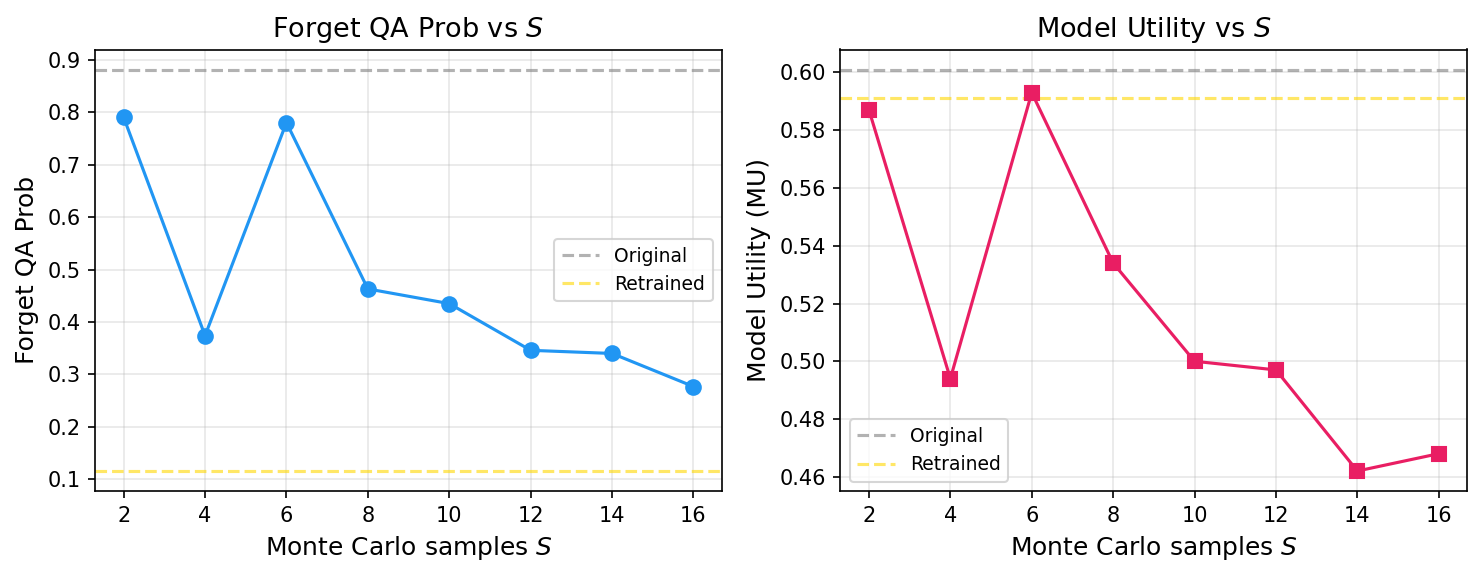}
\end{center}
\caption{WIN-U trade-off curves across varying MC sample sizes $S$ on TOFU forget10.}
\label{fig:mc_sweep}
\end{figure}

Figure \ref{fig:winu_tradeoff_steering} also serves as an ablation over step size $\eta$, showing that it provides a simple and efficient way to control the forget-retain trade-off of WIN-U.
We also conduct ablations on the number of MC samples $S$, which is shown in Figure \ref{fig:mc_sweep}.
In theory, as $S \to \infty$, the MC-WIN-U update should converge to the exact GGN-based WIN-U update, thus better forget performance. 
The results validate this trend, showing that increasing $S$ generally drops the forget QA probability. 
However, the sharp oscillation in the figure also reveals stochasticity and variance in the MC estimation, especially for smaller $S$ values.

\section{Extension to alternative unlearning objectives}
\label{app:extensions}

While matching the retraining optimum is the most principled definition of unlearning, certain applications may benefit from alternative objectives.
We show that the WIN-U framework extends naturally to two such settings.

\subsection{Maximizing forget loss while minimizing retain loss}

In this setting, the unlearning objective is a bi-objective optimization:
\begin{equation}
   \boldsymbol{\theta}_r^* = \arg\min_{\boldsymbol{\theta}}\; \mathcal{L}(\boldsymbol{\theta}) - (1+\gamma)\,\mathcal{L}_f(\boldsymbol{\theta}),
\end{equation}
where $\gamma > 0$ controls the forget--retain trade-off.
Following the same Newton--GGN--Woodbury derivation as in Section~\ref{sec:winu_update}, the update becomes:
\begin{equation}
   \boldsymbol{\theta}_r^* \approx \boldsymbol{\theta}^* + \frac{1+\gamma}{n}\mathbf{H}^{-1} \mathbf{J}_f^\top \left(\mathbf{I}_{mc}- \frac{1+\gamma}{n}\mathbf{B}_f \mathbf{J}_f \mathbf{H}^{-1} \mathbf{J}_f^\top\right)^{-1} \boldsymbol{\delta}_f.
\end{equation}

\subsection{Target output on forget set}

Another scenario is to redirect the model output on $\mathcal{D}_f$ toward a target value $\hat{y}_j$ (e.g., a random or average label):
\begin{equation}
   \boldsymbol{\theta}_r^* = \arg\min_{\boldsymbol{\theta}}\; \mathcal{L}(\boldsymbol{\theta}) - \mathcal{L}_f(\boldsymbol{\theta}) + \frac{\gamma}{n}\sum_{j=1}^{m} \ell(\boldsymbol{\theta}, \mathbf{x}_j, \hat{y}_j).
\end{equation}
The corresponding update is:
\begin{equation}
   \boldsymbol{\theta}_r^* \approx \boldsymbol{\theta}^* + \frac{1}{n}\mathbf{H}^{-1} \mathbf{J}_f^\top \left(\mathbf{I}_{mc}- \frac{1}{n}\mathbf{B}_f \mathbf{J}_f \mathbf{H}^{-1} \mathbf{J}_f^\top\right)^{-1} (\boldsymbol{\delta}_f - \gamma\,\hat{\boldsymbol{\delta}}_f),
\end{equation}
where $\hat{\boldsymbol{\delta}}_f$ is the stacked output-gradient vector evaluated at the target labels $\hat{y}_j$.

\end{document}